\documentclass[12pt,authoryear]{elsarticle2}
\usepackage{xr-hyper}
\usepackage[utf8]{inputenc}
\usepackage[T1]{fontenc}

\usepackage[usenames,dvipsnames]{xcolor,colortbl} 
\usepackage{moreverb,url}
\usepackage{caption}
\usepackage{subcaption}
\DeclareUnicodeCharacter{0301}{\'{e}}
\usepackage[colorlinks,bookmarksopen,bookmarksnumbered,citecolor=red,urlcolor=red]{hyperref}
\makeatletter
\def\blfootnote{\gdef\@thefnmark{}\@footnotetext}
\makeatother

\newcommand\BibTeX{{\rmfamily B\kern-.05em \textsc{i\kern-.025em b}\kern-.08em
T\kern-.1667em\lower.7ex\hbox{E}\kern-.125emX}}

\usepackage{tabularx}

\usepackage{efbox,graphicx}
\efboxsetup{linecolor=red,linewidth=3pt}
\usepackage{esvect}

\usepackage{xcite}

\makeatletter
\newcommand*{\addFileDependency}[1]{
  \typeout{(#1)}
  \@addtofilelist{#1}
  \IfFileExists{#1}{}{\typeout{No file #1.}}
}
\makeatother

\newcommand*{\myexternaldocument}[1]{
    \externaldocument{#1}
    \addFileDependency{#1.tex}
    \addFileDependency{#1.aux}
}

\usepackage[export]{adjustbox}
\usepackage[noend]{algpseudocode}
\usepackage{algorithm}

\usepackage{tikz}
\usepackage{amsmath,amssymb,amsthm}
\usepackage{dsfont}
\usepackage{graphicx}
 \usepackage{amsbsy}

\usepackage{textcomp}
\usepackage{siunitx}

\newcommand{\vx}{\vv{x}}
\DeclareMathOperator*{\argmax}{arg\,max}
\DeclareMathOperator*{\argmin}{arg\,min}

\myexternaldocument{SI}

\usepackage[symbol]{footmisc}
\begin{document}
\begin{frontmatter}

\title{Dealing with Expert Bias in Collective Decision-Making}

\author[1,2]{Axel Abels\corref{cor1}%
}
\ead{axel.abels@ulb.be}
\cortext[cor1]{Corresponding author}
\author[1,2,3]{Tom Lenaerts}
\ead{tom.lenaerts@ulb.be}
\author[4]{Vito Trianni}
\ead{vito.trianni@istc.cnr.it}
\author[2]{Ann Now{\'e}}
\ead{ann.nowe@vub.be}

\affiliation[1]{organization={Machine Learning Group, Universit{\'e} Libre de Bruxelles},
            addressline={Boulevard du Triomphe, CP 212},
            city={B-1050 Brussels},
            country={Belgium}
            }
\affiliation[2]{organization={AI Lab, Vrije Universiteit Brussel},
            addressline={Pleinlaan 2},
            city={B-1050 Brussels},
            country={Belgium}
          }
\affiliation[3]{organization={Center for Human-Compatible AI, UC Berkeley},
            addressline={2121 Berkeley Way},
            city={94720 Berkeley, CA},
            country={USA}
           }

\affiliation[4]{organization={Institute of Cognitive Sciences and Technologies, National Research Council},
            addressline={Via S. Martino della Battaglia, 44},
            city={00185 Roma (RM)},
            country={Italy}
           }

\begin{abstract}%
Quite some real-world problems can be formulated as decision-making problems wherein one must repeatedly make an appropriate choice from a set of alternatives. Multiple expert judgements, whether human or artificial, can help in taking correct decisions, especially when exploration of alternative solutions is costly. As expert opinions might deviate, the problem of finding the right alternative can be approached as a collective decision making problem (CDM) via aggregation of independent judgements. 
Current state-of-the-art approaches focus on efficiently finding the optimal expert, and thus perform poorly if all experts are not qualified or if they are overly biased, thereby potentially derailing the decision-making process. In this paper, we propose a new algorithmic approach based on contextual multi-armed bandit problems (CMAB) to identify and counteract such biased expertise. We explore homogeneous, heterogeneous and polarised expert groups and show that this approach is able to effectively exploit the collective expertise, outperforming state-of-the-art methods, especially when the quality of the provided expertise degrades. Our novel CMAB-inspired approach achieves a higher final performance and does so while converging more rapidly than previous adaptive algorithms. 
\end{abstract}


\end{frontmatter}

\section{Introduction}
\label{Introduction}

 The recent emergence of COVID-19 \citep{world2020coronavirus} has presented diverse groups of decision-makers with substantially the same problem: what measures to put in place to reduce the loss of human life while limiting economic decline and ensuring mental well-being? While the single best sequence of decisions is unlikely to be identical for any two regions, governments have taken strikingly different approaches to handle the pandemic \citep[e.g. the difference between Sweden and other EU countries, see][]{RePEc:hhs:lunewp:2020_011}. Whether due to legislative restrictions on the available decisions, cultural differences, or past experiences, governments---influenced by the information and expertise available to them---have not been uniform in their response to the pandemic \citep{anderson2020will}. It has become increasingly clear that experts' advice is not aligned, begging the question as to which advice should be followed. This problem is apparent especially in the current pandemic, as previous knowledge on how to manage a pandemic cannot be reliably exploited due to the specificity of the virus and the variability of the socio-economical context. What followed instead was a series of regulation adjustments as governments re-evaluated previously taken measures in terms of their effectiveness and the limitations they imposed, weighing them in the light of the advice provided by experts and the demands coming from the electorate and different lobbies. Biases displayed by experts may be many, as for instance different goals (e.g., a rapid economic recovery), different levels of knowledge about epidemics, level of exposure to a health crisis of this scale or even personalities. It is well known that, when biases are present, the effectiveness of deciding in a group is reduced, sometimes leading to major mistakes \citep{bang2017making}. Hence, methods to identify and counteract possible biases are paramount.
 
 In this paper, we take a computational stance to address the problem of taking appropriate decisions in a group of experts (hereafter referred to as collective decision making---CDM) in an online setting, explicitly accounting for lacking expertise. Specifically, we propose an algorithmic solution that finds the best options from a set of alternatives, exploiting in the best possible way the advice from biased experts.

We consider the setting wherein experts --- whether humans or AI algorithms --- observe the problem's description and provide advice about the alternative solutions. In other words, each expert proposes a solution to the decision problem, in the form of advice. This advice can then be exploited centrally in conjunction with other experts' advice to determine the best possible alternative, e.g., through some voting process or other decision making approaches. A desirable outcome in such scenarios is that of collective intelligence, which occurs when the solution obtained through the collective surpasses the single best solution in the collective. For example, when it is unclear which AI algorithm is most suited to the problem at hand, the expert framework allows an artificial learner to identify the best way to act on the advice of artificial experts, thus enhancing performance through a collective of artificial intelligences. While a simple majority vote can lead to improvements in performance under certain assumptions \citep[see Condorcet's jury theorem, ][]{nicolas1785essai}, more complex  methods have also been proposed. 
For instance, algorithms for \emph{bandits with expert advice} \citep{doi:10.1137/S0097539701398375} learn to weigh advice of good experts more strongly. Previous work has approached this CDM problem with methods that are focused on efficiently identifying the best available expert (i.e., no solution is considered outside the set of solutions provided by the experts). In such cases, performance is measured relative to the best expert \citep{doi:10.1137/S0097539701398375,DBLP:journals/corr/Zhou15c,agarwal2012contextual,Foster2018}. When this single best expert is optimal --- which is a standard assumption when considering a set of regression experts \citep{agarwal2012contextual,foster2020beyond}---, these methods efficiently allow a learner to converge towards optimality. 
However, for these methods, poor absolute performance becomes inevitable when the set of experts is itself limited in terms of performance, which calls for different solutions. In this work we posit that, by combining the advice of experts, as opposed to selecting a single expert, we can perform better than the single best expert in most scenarios.  

When learning how to act on the advice of experts in an online setting, differences in expertise induce high uncertainty about the outcomes of choosing one or the other alternative, as limited or no knowledge of expert competence is available. It is therefore crucial to take an approach which carefully balances exploration---to reduce this uncertainty---and exploitation---to optimize the outcome of our decisions.
To address this uncertainty, we build on existing work on bandit problems in order to maximally exploit the potential of the expert set. In particular, we propose a novel approach to CDM which reduces the problem of CDM in an online setting to a contextual multi-armed bandit (CMAB). We refer to this approach as Meta-CMAB. 
Starting from a formalization of CDM as a CMAB problem, we carefully evaluate the applicability of various exploration principles. We then investigate in this paper how the Meta-CMAB approach overcomes systematic biases in expert advice. In particular, we study the effect of varying degrees of homogeneity and polarization in the expert population. We will show that Meta-CMAB is robust to such biases as long as expert advice is (positively or negatively) correlated with the actual outcome, in which case its performance often surpasses the single best expert. Furthermore, the Meta-CMAB also takes advantage of the presence of under-performing experts, who can often still  provide relevant contributions to the decision-making process if they perform worse than random. We contrast these results with the state-of-the-art approaches to deal with decision making with expert advice, namely a reduction to a MAB, as introduced by \citet{auer2002finite}, and a more sophisticated approach which optimally converges towards the single best expert, EXP4-IX \citep{neu2015explore}. When taking their set of policies or regressors to be the expert set, these algorithms converge towards the single best expert's policy, limiting the opportunities for collective intelligence. 

Before introducing our collective intelligence enabling method, Meta-CMAB, and its relation to previous methods, we first present the necessary background knowledge on CMABs and bandits with expert advice. We then provide a theoretical analysis of the performance of Meta-CMAB, taking into account characteristics of the expert set such as homogeneity and polarization. Finally, we perform an extensive experimental comparison of Meta-CMAB and previous state-of-the-art algorithms, on both synthetic and human expert data sets.  

\section{Background}
\label{background}

CDM \citep{aikenhead1985collective,ROBSON2010230,MARSHALL2017636} is concerned with combining the opinions of a group of experts to make appropriate decisions. More specifically, we are interested in using a group of experts to maximize the decision-making accuracy throughout a sequence of consistent problem instances.
In medical diagnostics for example, each decision in a sequence coincides with the choice of a treatment to be administered to a patient. Given this sequence, a rational goal is to maximize the number of correct diagnoses and appropriate treatments administered with support from a group of expert diagnosticians \citep{10.1001/jamanetworkopen.2019.0096}. Similarly, in a pandemic, taking into account the demographics of a population, the estimated basic reproduction number \citep[$R_0$, ][]{milligan2015essential}, the number of infections, and the current capacity for hospitalization, one might implement more or less extreme measures in the hope of containing the virus while minimizing economic recession. When assessing the situation for a different region,  a decision-maker might come to a different decision. 
As a result of a lack of certainty about the potential outcomes, sub-optimal choices are often inevitable but necessary to enhance a learner's understanding of the consequences associated with different choices. This fundamental dilemma between the need for expanding understanding and the desire to make the best decisions is formalized by MABs, which we introduce in the following section. We then present prior research on solving CDM problems formalized as bandits with expert advice. Tackling this formalization is the focus of our Meta-CMAB approach, which we present in \autoref{sec:metacmab}.

\subsection{Multi-Armed Bandits}

A MAB is characterized by a set of $K$ arms identified by numbers $1,...,K$ and a value mapping $f: [K] \rightarrow [0,1]$ from an arm to a real-valued expected reward \citep{agrawal1995sample}. In the context of decision-making as we discussed in the introduction, each arm is an alternative we can choose, for example a treatment. When choosing an arm $k$, we observe a noisy reward $r$ sampled from a distribution with mean $f(k)$. Because $f$ is unknown, exploration is required to identify the best arm(s). State-of-the-art MAB algorithms balance exploration of uncertain arms and exploitation of likely optimal arms to maximize the sum of rewards \citep{russo2017tutorial}.
An agent's policy at time $t$, $\pi_t \in \Delta([K])$, with $\Delta([K])$ the $K$-dimensional probability simplex, is a probability distribution over the arms according to which the agent chooses an arm $k_t \sim \pi_t$. Because the optimal performance we can expect to attain is that of the best arm, performance is often measured in terms of regret. That is, the difference between the cumulative value of the best policy (which always chooses the arm maximizing $f$) and the cumulative value of the learner's policies. Let $(k_t)_{t=1}^T$ be the sequence of arms pulled by the learner, we define the regret as
$$R_T = \sum_{t=1}^T\big( \max_{k\in [K]} f(k) - f(k_t) \big)$$

The aim of any agent is to minimize the regret $R_T$, or, equivalently, maximize the sum of collected rewards. 

While MABs are useful as models of fundamental repeated decision-making, they are limited in their applicability as they assume that arms are only characterized by their identifier. 
In real world problems, decisions are typically made based on additional information about the problem or the arms. In medical diagnostics for example, it is necessary to take into account characteristics specific to the patients, such as their medical record. The addition of such information transforms a MAB into a contextual MAB (or CMAB).

\subsection{Contextual Multi-Armed Bandits}

In contextual bandits \citep{auer2002using,wang2005bandit,chu2011contextual}, each decision (or time step) is characterized by a $d$-dimensional context vector $\vx_t$ sampled from some unknown distribution, and rewards are determined by a fixed but a-priori unknown mapping of an arm context vector to a value, i.e., a function $f: [K] \times \mathds{R}^d \rightarrow [0,1]$. Similar to the simple MAB, when pulling an arm $k$ we observe a reward sampled from some distribution with mean $ f(k,\vx_{t})$. In medical diagnostics for example, the context vector could be a set of features describing the concerned patient (e.g., their medical record), and each possible treatment (arm) affects the mapping from this context to an outcome, e.g., whether the treatment was successful, or a QALY score \citep{whitehead2010health}.
In contextual bandits, an agent's policy, $\pi : \mathds{R}^{d} \rightarrow \Delta([K])$, maps the context vector onto a probability distribution over the arms.

Similarly to the MAB case, we define the regret of a policy selecting the sequence of arms $(k_t)_{t=1}^T$ as:
\begin{equation}
\label{eq:cmab_regret}
 R_T = \sum_{t=1}^T \big( \max_{k\in [K]} f(k,\vx_{t}) -  f(k_t,\vx_{t}) \big ) 
 \end{equation}

When approximating $f$ from scratch is impractical, experts can provide the advice required to select the appropriate arms. 
Identifying the best way to use this advice can however present another challenge, as all experts might not be equal in their expertise. Both static aggregation techniques such as plurality votes \citep{Grofman1983}, or adaptive methods which for example learn to identify the best performing experts \citep{ doi:10.1137/S0097539701398375, beygelzimer2011contextual} have been proposed to address this. 
In the following section we formalize this problem of bandits with expert advice and present different algorithmic approaches to tackle it. In addition we detail how previous algorithms in similar settings differ from ours, and why our new approach is relevant.  

\subsection{Bandits with Expert Advice}\label{sec:DwEA}
 The problem of bandits with expert advice \citep{doi:10.1137/S0097539701398375,agarwal2012contextual} formalises a CDM process wherein at each time step $t$ a set of $N$ experts observe the context of a CMAB and provide the centralized learner with advice based on their prior knowledge about the problem. This knowledge for each expert $n$ is captured by an approximation of $f$, i.e. $\tilde{f}^n: [K] \times \mathds{R}^d \rightarrow [0,1]$. 
 In other words, given the problem description $\vv{x}_t$ observed at time $t$, expert $n$ is a regressor whose prediction of arm $k$'s expected reward is $\tilde{f}^n(k,\vv{x}_t)$. 

  
 We make no assumptions on how experts acquire their expertise. In case of human experts, it can for example be the result of past experiences; for an AI expert, it can be acquired by training on an existing (possibly biased) data set. 
 In any case, the aim of the centralized learner in this setting is to use the approximations of $f$ provided by the expert set to act optimally in the underlying bandit. 
 
 For brevity, we will denote the advice of expert $n$ for arm $k$ at time $t$, $\tilde{f}^n(k,{\vv{x}_t})$ as $\tilde{f}^n_{k,t}$. 
 We further construct expert $n$'s advice vector induced by the context at time $t$ as $\vv{\tilde{f}}^n_t=[ \tilde{f}^n_{1,t},...,\tilde{f}^n_{K,t}]$, and let
 $\pmb{\tilde{f}}_t=[\vv{\tilde{f}}^1_t \cdots \vv{\tilde{f}}^N_t]^\intercal$ be the advice matrix at time $t$, wherein row $n$ thus consists of expert $n$'s advice vector.

 A learner for bandits with  expert advice selects arms based on this advice matrix by maintaining a time-dependent policy $\pi_t$ which maps expert advice to a probability distribution over the arms. By incorporating experiences (i.e., chosen arms, the advice that led to each chosen arm, and the observed rewards), the learner should improve its policy while, again, balancing exploration and exploitation in order to maximize the cumulative reward, or equivalently, minimize regret. \autoref{alg:expert-learning} outlines the problem of bandits with expert advice. An important observation here is that the learner is context-agnostic. By this we mean that only the experts need to observe the context, the learner itself does not explicitly take the context into account to make its decision. As a result, it is also applicable to problems for which the contexts can be fully observed by human experts but are hard to meaningfully capture into data points to train a CMAB solver.
 
\begin{algorithm}
 \caption{Bandits with expert advice}\label{alg:expert-learning}
 \begin{algorithmic}[1]
 \Require underlying contextual bandit with reward function $f: [K] \times \mathds{R}^d \rightarrow [0,1]$, and $N$ experts providing approximations $\{\tilde{f}^n\}_{n=1}^N$
 \State Initialize learner with initial policy ${\pi}_1$
 \For{$t = 1, 2, ..., T_{cdm}$}
 \State Experts observe the context ${\vv x}_t$
 \State {Get expert advice matrix  $\pmb{ \tilde{f}}_t=[\vv{\tilde{f}}^1_t, ..., \vv{\tilde{f}}^N_t]^\intercal$} %
 \State Pull arm $k_{t} \sim \pi_t(\pmb{ \tilde{f}}_{t})$ and collect resulting reward $r_t$\label{step:expert-learning-policy}
 \State $\pi_{t+1} = update(\pi_t,\pmb{ \tilde{f}}_{t}, k_t,r_t)$ \label{step:expert-learning-update}
 \EndFor
 \end{algorithmic}
 \end{algorithm}

Performance is typically measured in terms of regret with respect to the single best expert. Algorithms thus compete with the single best expert in hindsight. Let $k_{t}^{\tilde{f}}=\argmax_{k\in[K]} \tilde{f}(k,\vx_t)$ be the arm chosen at time $t$ by greedily acting on some approximation $\tilde{f}$, and let $(k_t)_{t=1}^T$ be the sequence of actions taken by the learner, the regret is then

\begin{equation}
\label{eq:bwea_regret}
 R_T = \max_{n \in [N]} \sum_{t=1}^T \big(  f(k_{t}^{\tilde{f}_n},\vx_{t}) -  f(k_t,\vx_{t}) \big ) 
 \end{equation}

This regret expresses the natural goal of performing as well or better than the single best expert. Under the realizability assumption (there exists an expert $n$ such that $\tilde{f}^n = f$), this regret is equivalent to the regret with regards to optimality. When the realizability assumption does not hold, i.e., when only sub-optimal experts are available, this regret does not give us a clear indication of performance on the underlying CMAB. 
In such a case, the regret as defined in \autoref{eq:cmab_regret} for contextual bandits is more indicative of performance relative to optimality, irrespective of the quality of the expert set.  

Whether it is the result of inaccurate past experiences, cognitive biases \citep[e.g., outcome bias, see][]{baron1988outcome}, or simply malicious intent \citep[which can entail that the expert's values are not aligned with those of the collective, possibly as a result of reactance,][]{steindl2015understanding}, biases are likely to be introduced along the knowledge acquirement process \citep{o2018cognitive,dror2018expert} and manifest themselves in the expert's advice.

This work focuses on maximizing the potential of small, imperfect, classes of regressors, e.g., those we would obtain when consulting humans. In particular, given an imperfect set of experts, we explore whether it is possible to act on the imperfect set in such a way that we outperform the single best expert, thus attaining \emph{collective intelligence}. Clearly this can only occur if the realizability assumption does not hold, i.e., if the best expert is not optimal, a likely case in real world settings.

Despite this likely sub-optimality, the focus of existing work is on identifying the single best expert. To illustrate this, we discuss some state-of-the-art approaches for bandits with expert advice in the following section.

\subsubsection{Baseline algorithms}
A straightforward approach in collective decision-making is to act greedily on the averaged advice of experts. 
 This method is particularly effective when the performance of experts is similar and better than random.  Experts however are not necessarily equal in performance, and additionally can show varying degrees of correlation among themselves, two factors which degrade the performance of the simple average. If knowledge about the experts' expected performance is available (e.g., a confidence estimate), votes can be weighted in function of this performance estimate, resulting in a weighted majority vote \citep[WMV, see][]{MARSHALL2017636} which weighs each expert's advice by its confidence to obtain a value for each arm and then greedily acts on the aggregate.

When confidence estimates are inaccurate, the performance of a WMV can quickly degrade \citep{abels2020exp4con}. It can therefore be desirable to independently approximate performance estimates when confidence estimates are unavailable or inaccurate.

In alternative to acting on an averaged advice, a learner can try to identify and act on only the best expert in the pool. To this end, it is straightforward to draw a parallel between the previously introduced MAB problem and the problem of bandits with expert advice, as proposed by \cite{doi:10.1137/S0097539701398375}. In essence, given $N$ experts, each expert can be treated as an arm in an $N$-armed meta-MAB. Selecting an "arm" means we only follow the corresponding expert's advice. Concretely, if at time $t$ arm $n$ of the meta-bandit is pulled, we greedily pull arm $k_{t}^{\tilde{f}^n}=\argmax_{k\in[K]} \tilde{f}^n(k,\vx_t)$ in the original bandit. The "arm"'s estimated value is then updated based on the reward observed by following its advice.

The Meta-MAB is of particular interest for two reasons. Conceptually it is a straightforward algorithm which provides interpretable estimates of expert quality. In addition, in a setting wherein querying multiple experts is costly, Meta-MAB minimizes the number of experts to be queried at each round to a single expert. This is not the case for the algorithms that follow, and we therefore include Meta-MAB as a baseline algorithm. Supplementary \autoref{sec:metamab} provides a full description of the Meta-MAB algorithm.
In spite of these benefits, as expert advice is never combined, this approach's performance is bounded by the performance of the best expert. What is more, at each time step only the estimated value of one expert is updated. However, as experts provide advice on the same problem, observed rewards could be used to update estimates about all experts. When the expert set is relatively large, these simultaneous updates enable better performance. 

 The exponential weighting method EXP4-IX \citep{auer2002finite,neu2015explore} for example, maintains a weight distribution over the experts, which it updates in function of observed rewards in order to iteratively increase the importance of experts whose advice enables high rewards. While it is more efficient in the sense that it uses each observation to update beliefs about all experts, EXP4-IX cannot enable collective intelligence when its set of policies is the set of expert's greedy policies. A full description of the EXP4-IX algorithm is provided in supplementary \autoref{sec:exp4pcon}. Among algorithms that compete with the single best policy, EXP4-IX has optimal theoretical guarantees and is straightforward to implement. For large (possibly infinite) expert sets the computational cost of EXP4-IX can however be prohibitive. In such cases, alternatives to EXP4-IX are required. Algorithms adapted to large (possibly infinite) expert sets such as the \emph{Epoch-Greedy} algorithm \citep{langford2007epoch} or \emph{ILOVETOCONBANDITS} \citep[ILTCB for short,][]{agarwal2014taming} sacrifice some performance and generality in order to reduce computational cost. In the natural setting wherein the set of policies over which they optimize is the set of experts, these methods are similarly restricted to the performance of the single best expert. 
Given our focus on smaller expert sets, the additional computational complexity of EXP4-IX is negligible and we therefore select it for inclusion as a reference algorithm for policy selection.
\paragraph{Inversion Method}
 For larger biases, the presence of negatively correlated experts presents a challenge to methods utilizing a simple weighted average (notably Meta-MAB and EXP4-IX). In such a model, the optimal weight of worse-than-random experts is $0$, implying that the knowledge present in bad advice (intuitively, which actions to avoid), is not exploited. A simple approach to potentially improve the performance of weighted average algorithms in the presence of worse-than-random experts is to augment the expert set with inverted duplicates of available experts. In doing so, the knowledge of negatively correlated experts can be utilized by meta-MAB or EXP4-IX for example.

Note that, while we consider here experts in the form of regressors, policy-based methods such as EXP4-IX can be applied on the set of $N$ policies induced by greedily acting on each of the experts' advice. This transformation from a regressor set to a policy set induces some loss of information, which suggests a more appropriate approach is to directly optimize over the set of regressors. 

One approach to optimizing directly over the set of regressors is to iteratively restrict the set of regressors, and to act only on the advice of the restricted set. This ensures that only regressors with small prediction errors are followed, which leads to appropriate arm choices. Regressor Elimination \citep{agarwal2012contextual} or RegCB.elimination \citep{Foster2018} for example follow this paradigm. Note however that these algorithms rely on the realizability assumption to ensure the single best expert is preserved with high probability. When this assumption does not hold, these algorithms can fail to converge towards the single best expert. Note also that, in the natural case wherein we apply these algorithms on the set of experts, a convergence towards a single expert necessarily excludes the possibility for collective intelligence. 

In supervised learning the aggregation of multiple regressors has been explored before \citep{breiman1996stacked}. However, when applied to the bandit setting, this method fails to provide the necessary exploration. Similarly, online gradient boosting \citep{beygelzimer2015online} learns to combine several weak regressors to output good predictions in online regression. However, their setting differs in that it is not a bandit setting, thus the problems induced by bandit feedback are not considered. The work on boosting of \cite{hazan2021boosting} does consider the bandit setting, but differs from our setting in how experts interact with the centralized learner. Experts in these boosting approaches are controlled by the learner, as they learn through data and feedback provided by the learner. Rather than learning to weigh experts, online gradient boosting learns to train experts. It follows that if the experts are not trainable (or at least do not incorporate new experiences as expected by these boosting approaches) as is the case for human experts or AI models which are not under control of the learner, this approach is not applicable. However, even when experts match the assumptions imposed by this approach, the resulting bandit algorithm results in performance with significantly weaker theoretical guarantees than the approaches we consider in this work.

\subsubsection{Oracle approaches}
As an alternative to the previously discussed algorithms, a vein of algorithms make use of an oracle in order to optimize performance, see for example SquareCB \citep{foster2020beyond}, FALCON \citep{Simchi-Levi2020} or ILTCB \citep{agarwal2014taming}. While the assumptions imposed on the oracle can change, these algorithms rely on a black-box-like oracle to return the best in class policy/regressor. Typically this class is expected to be a hypothesis space, such as the set of linear regressors, and the oracle is a regression algorithm from which the empirically optimal function can be efficiently queried. When applied over our set of base experts, this optimal function would be the (likely imperfect) best expert. If applied this way, these oracle methods thus display the same drawbacks as EXP4-IX in terms of their inability to surpass the best expert.

\

The different methods presented so far provide a gamut of approaches to identifying the best expert while providing sufficient exploration. It should however be highlighted that any such method, when simply applied to our limited set of experts would be similarly limited.

Instead, and with the aim of outperforming the single best expert, our primary contribution is an approach which converges towards the best aggregation over the expert set. 

\section{Meta-CMAB: meta-learning for Bandits with Expert Advice}\label{sec:metacmab}
It has long been understood that the knowledge of collectives can, under the right conditions, be aggregated to surpass even the best individual in the collective \citep[see for example Condorcet's jury theorem, ][]{nicolas1785essai}. Simple averages for example have been successfully applied as a basic aggregation technique \citep{Kattan2016,kammer2017potential}. Unfortunately, among all linear combinations of advice, simple averages are optimal only when experts are i) better than average, ii) have similar levels of expertise, and iii) correlation among experts is uniform. While there are some settings in which these conditions can be expected to approximately hold, our desire is to design an approach which is robust when they do not hold.

Our primary contribution is to frame the task of optimizing performance from a set of experts as a contextual bandit, and exploring the applicability of existing algorithms to this particular case. 

Fundamentally, the aim of our Meta-CMAB approach is to efficiently search large hypothesis spaces for an optimal mapping from advice.  
Specifically, we make the assumption that there exists some function $\mathcal{E}$ which maps the experts' advice to an expected reward, such that:
$$ \mathds{E} (r_{k,t}) = f(k,\vx_t) = \mathcal{E}(\vv{\tilde{f}}_{k,t}) $$
If such an $\mathcal{E}$ exists, we can construct a secondary disjoint CMAB, the Meta-CMAB, with arm context 
$\vv{c}_{k,t}= \{\tilde{f}_{k,t}^1,...,\tilde{f}_{k,t}^N\}$ for each arm $k$. Minimizing regret in this Meta-CMAB also minimizes regret in the base bandit.  
Conceptually, the consequence in terms of regret is that we do not compete with the single best expert, but rather with the best combination of experts. 

 This reduction makes it possible to select an appropriate CMAB algorithm to solve the Meta-CMAB, and consequently the original problem. Just as in standard CMABs, the choice of the CMAB algorithm depends on the assumptions made about $\mathcal{E}$.
 
 
 \begin{algorithm}
 \begin{algorithmic}[1]
 
 \Require underlying contextual bandit with reward function $f: [K] \times \mathds{R}^d \rightarrow [0,1]$, and $N$ experts providing approximations $\{\tilde{f}^n\}_{n=1}^N$
 \State Initialize a CMAB algorithm with initial policy ${\pi}_1$
 \For{$t = 1, 2, ..., T$}
 \State Experts observe the context ${\vv x}_t$ 
 \State {Get expert advice $\pmb{ \tilde{f}}_t=\{\vv{\tilde{f}}^1_t, ..., \vv{\tilde{f}}^N_t\}$ } \label{linucb:outline:advice}
 \State ${\vv {\tilde{f}}}_{k,t} = \{\tilde{f}_{k,t}^1,...,\tilde{f}_{k,t}^N\}\; \forall k\in\{1,...,K\}$ \Comment construct meta-contexts
 \State Pull arm $k_{t} \sim {\pi}_t(\{{\vv {\tilde{f}}}_{1,t},...,{\vv {\tilde{f}}}_{K,t}\})$ and collect resulting reward $r_t$ \label{linucb:outline:arm}
 \State ${\pi}_{t+1} = update({\pi}_t,{\vv {\tilde{f}}}_{k_t,t}, k_t,r_t)$ 
 %
 \EndFor
 \end{algorithmic}
 \caption{Meta-CMAB for bandits with expert advice}\label{alg:value-pred}
 \end{algorithm}
With the aim of providing an interpretable approach, we focus here on the linear case. In particular, we aim to identify the best linear combination of expert advice. That is, the weight vector
\begin{equation} 
\vv{\theta^*} = \argmin_{\vv{\theta} \in \mathds{R}^N} \mathds{E} \large[(f(k,\vx) - \langle  \vv{\tilde{f}_k}, \vv{\theta}  \rangle)^2 \large]\end{equation}\label{eq:optimal_weights}
If there exists a weight vector such that the expectation above is $0$, greedily acting upon the linear combination leads to $0$ regret w.r.t. optimality. When no such weight vector exists, we say the model is misspecified, i.e., there is no $\vv{\theta^*}$ such that $ f(k,\vx_t) =  \mathds{E} [\langle \vv{\tilde{f}_{k,t}}, \vv{\theta^*} \rangle]$.

Note that while one could simply act greedily on a linear regression trained on the experience history, the bandit setting requires us to be more careful in our decision-making.

 Several approaches have been proposed to provide the necessary exploration in contextual bandits.
 In what follows we explore the viability of common principles used to solve contextual bandits and discuss their implications when used to solve the constructed meta-contextual bandit whose features consists of expert advice. 

\subsection{Inverse Gap Exploration}
As we discussed previously, FALCON relies on an oracle to query the best known regressor. While an intuitive application to our problem would see the set of regressors be exactly our expert set, we propose to consider a larger regressor class. Specifically, we consider the class of linear combinations of expert advice. At each time step, FALCON would then act on the empirically optimal linear combination, i.e., the estimates $\hat{f}_{k,t} = \langle \vv{\tilde{f}_{k,t}}, \vv{\theta} \rangle $. FALCON then uses \emph{inverse gap weighting} to derive a probability distribution over the actions which concentrates more mass on the arms with high estimated reward. 
Hence in practice, the larger the gap between the estimates of arms $k$ and the estimated best arm, the lower the probability of pulling arm $k$. This means that exploration is not explicitly driven by the expert advice. Exploration is only driven by the estimate differences, regardless of possibly differing estimate variances. In order to appropriately drive exploration, FALCON assumes the model is not misspecified, which in our case implies $f$ can be perfectly estimated by a linear mapping of expert advice. When expertise is imperfect however, this linear model is likely misspecified.

To handle such misspecification, SafeFALCON \citep{Krishnamurthy2021} introduces a statistical test which interrupts training when misspecification is detected and then proceeds with the best known model so far. This test consists in checking whether performance has started to degrade, a sign of misspecification. Using this technique, the authors show that performance gracefully degrades with misspecification. 
We refer to the method which constructs the Meta-CMAB and uses Safe-FALCON to optimize over it as \emph{Meta-CMAB (FALCON)}.



\subsection{Learning To Explore}
ILTCB \citep{agarwal2014taming}, as discussed before, is an alternative to EXP4-IX which aims to efficiently optimize over large policy sets. While it is not advantageous to apply it over smaller expert sets, we now consider an enhanced expert set. Specifically, we propose to feed the class of linear aggregations of expert advice into ILTCB. In other words, for each possible weight vector $ {\vv\theta} \in \mathds{R}^N$, we include a policy which acts greedily on the estimates $\langle {\vv\theta} , {\vv {\tilde{f}}_{k,t}} \rangle$. Note that while this policy set is infinite, we can efficiently query policies through oracles as detailed in \cite{agarwal2014taming}. While similar to SquareCB or FALCON, in that it uses oracle calls to query suitable policies, it differs from these by maintaining a (sparse) distribution over the policies. This distribution is constructed in a way that balances exploration and exploitation. 
Note that ILTCB makes no assumption of realizability, thus it can be applied as is to the class of linear policies over the advice. 
ILTCB does however assumes finite policy classes, which the set of linear policies over the expert set is not. While we could approximate the class of linear policies by a cover, its size would be exponential in the number of experts, which would render ILTCB impractical. In particular, ILTCB performs uniform exploration which increases polylogarithmically in the size of the policy set. When the policy set grows exponentially with the number of experts, the amount of exploration performed by ILTCB becomes impractical. Dismissing the recommended exploration rate in favour of a lower value induces an excessive number of oracle calls, which renders this choice impractical. Considering this limitation, and to avoid the cost of offline oracles, the authors of ILTCB propose an online approximation of ILTCB called Online Cover. Like ILTCB, Online Cover maintains a sparse distribution over policies, but unlike ILTCB, it explicitly bounds the size of the distribution's support. In addition it differs in that it uses online oracles. Applying this algorithm to our problem allows us to learn to explore, as the distribution over policies learned by Online Cover should ensure that we are sufficiently exploitative, while also ensuring the necessary exploration to maintain an accurate model over the expert advice. When used to solve the constructed Meta-CMAB, we refer to this approach as \emph{Meta-CMAB (ILTCB)}.

\

While ILTCB implicitly learns to explore through the distribution over policies it maintains, tractable uncertainty measures are known for several models. For example, it is possible to estimate the expected variance of estimates acquired through linear regression. Exploring options whose variance estimate is high is a principle that has been successfully applied in contextual bandits. This principle, and its applicability to our setting are further discussed in the following.  

 \subsection{Using Advice to Drive Exploration} 
 Assume the reward equals the dot product between some ideal weight $\theta^*$ and the context vector --- here the advice vector --- plus some noise $\eta$ sampled from a $0$-mean, $\sigma$-subgaussian distribution:
 
 $$ r_{k,t} = f(k,\vx_t) + \eta =  \langle \vv{\tilde{f}_{k,t}}, \theta^* \rangle + \eta $$
 
 Applying ridge regression, we can approximate $\theta^*$ after $t$ timesteps as follows.
 Let $\vv{\tilde{f}}_{k_s,s}$ be the advice of the arm $k_s$ chosen at time $s$, and $r_{k_s}$ be the observed reward. 
 Let $\mathbf{V}_t = \lambda \mathbf{I} + \sum_{s=1}^t \vv{\tilde{f}}_{k_s,s} \vv{\tilde{f}}_{k_s,s}^\intercal$ be the correlation matrix and let $Y_t = \sum_{s=1}^t r_s\vv{\tilde{f}}_{k_s,s} $. 
The (regularized) weights minimizing \autoref{eq:optimal_weights} are $\hat{\theta}_t = \mathbf{V}_t^{-1} Y_t$. 
 Given these parameters, we can estimate the outcome of an arm as a function of the advice:  $$\hat{f}_{k,t} = \langle \hat{\theta}_t, \vv{\tilde{f}}_{k,t} \rangle $$
 The expected variance of this estimate is $\hat{\sigma}_{k,t}^2=\vv{\tilde{f}}_{k,t}^\intercal V_t^{-1} \vv{\tilde{f}}_{k,t} $ \citep{Li2010}. This variance thus gives us an indication of how large the error of our estimates is expected to be. With this knowledge we can make an appropriate choice of arm in order to maximize reward while also reducing uncertainty (characterized by high variance).
 
Given a history of experiences, this variance term is large for contexts, i.e., advice, which have rarely been seen before. In particular, this means that the use of this variance estimate to drive exploration, when applied on expert advice, provides increased exploration when experts (dis)agree in ways they usually do not. Such a change in the agreement of experts can for example indicate that some experts have acquired some knowledge independently, which can potentially lead to improved performance. This exploratory drive thus implicitly captures how we expect experts to behave with regards to each other, and ensures sufficient exploration to adapt when they change in unexpected ways. 

It remains unclear whether the noise model is appropriate for the expert setting. 
In particular, in addition to a noisy reward signal, it is likely that experts, especially humans, provide noisy advice.
Thus, instead of assuming that $$ r_{k,t} = \langle \vv{\tilde{f}_{k,t}}, \theta^* \rangle + \eta $$ we only assume that this holds with expectation, i.e., let $\vv{\eta_{\tilde{f}}}$ be a noise term on the advice of experts, $$ r_{k,t} =  \mathds{E}[\langle \vv{\tilde{f}_{k,t}}, \theta^* \rangle] + \eta =  \langle \vv{\tilde{f}_{k,t}} + \vv{\eta_{\tilde{f}}}, \theta^* \rangle + \eta_r =  \langle \vv{\tilde{f}_{k,t}}, \theta^* \rangle + \langle \vv{\eta_{\tilde{f}}}, \theta^* \rangle + \eta_r $$
Assuming $\eta = \langle \vv{\eta_{\tilde{f}}}, \theta^* \rangle + \eta_r $ is centered and conditionally sub-gaussian, this noise
model is equivalent. 

Aside from the noise, it is possible that the realizability assumption does not hold. 
Methods which do not account for this misspecification can fail to converge towards the best linear model \citep{Ghosh2017}. \citet{takemura2021parameter} and \cite{vial2022improved} propose Sup-Lin-UCB-Var, an algorithm designed with misspecification in mind. It uses the estimated variance to determine whether exploration is necessary. Specifically, at each timestep it selects the arms with maximal variance if this variance estimate is above a certain threshold. If no arm's variance is above the threshold, it is likely that estimates are accurate enough, in which case the learner simply acts greedily on its reward estimates. We refer to this approach as \emph{Meta-CMAB (UCB)}.


\paragraph{Regularization}
 In the equations above, $\lambda$ serves as a regularization term. This regularization can be particularly beneficial in the expert setting, as multi-colinearity is likely to occur when similar experts are present.

\section{Theoretical Analysis}\label{sec:theoretical}
We provide here bounds on the performance of Meta-CMAB and compare them to bounds obtained for the previously presented methods. 

Because the realizability assumption is unlikely to hold in real-world problems, there is a gap between the performance of the best expert available and that of an oracle.
Specifically, following previous work \citep{vial2022improved,Krishnamurthy2021}, we define misspecification as the largest estimate error over all contexts and arms:
$$ \varepsilon = \min_{\vv{\theta} \in \mathds{R}}  \max_{k\in [K]} \max_{t=1}^T | f(k,\vx_t) - \langle \vv{\theta} , \tilde{f}_{k,t}  \rangle| $$

Let us also define the expert misspecification, which is similar to the misspecification above, but restricted to the expert set:
$$ \zeta = \min_{n\in[N]}  \max_{k\in [K]} \max_{t=1}^T | f(k,\vx_t) - \tilde{f}^n(k,\vx_t) | $$

 Hence, over $T$ time steps, the best expert incurs a penalty of $O(\zeta T)$ relative to optimality.
Note that it is always the case that $\varepsilon \leq \zeta$. We study the gap between these two measures of misspecification in \autoref{sec:misspecification}.

\autoref{tab:regret} provides an overview of theoretical guarantees for the considered algorithms, which are explained briefly in the following subsections. 
\begin{table}[]
 \centering
 \begin{tabularx}{1.05\textwidth}{|>{}c | >{}c | >{}c |  }
 \hline 
 Approach &  Upper bound & References \\
 \hline
 MAB reduction & $O(\sqrt{TNlogT} + \zeta T)$ & \multicolumn{1}{p{2.715cm}|}{\citet{lattimore2020bandit}} \\
 EXP4-IX  & $O(\sqrt{KTlogN} + \zeta T)$ & \multicolumn{1}{p{2.715cm}|}{\citet{neu2015explore}} \\
 Meta-CMAB (UCB) & $\tilde{O}(\sqrt{NTlog(K)} + \varepsilon \sqrt{N}T)$ & \multicolumn{1}{p{2.715cm}|}{\citet{takemura2021parameter}} \\
 Meta-CMAB (FALCON)  & $O(\sqrt{KT(N+log(T))} + \varepsilon \sqrt{K}T)$ & \multicolumn{1}{p{2.715cm}|}{\citet{Krishnamurthy2021}} \\
 \hline 
 \end{tabularx}
 \caption{(Best known) theoretical bounds on different approaches.}
 \label{tab:regret}
\end{table}

\subsection{Reduction to MAB}
While the reduction to a MAB is straightforward, several algorithms to subsequently solve the MAB are available. The Bayesian regret of Thompson Sampling \citeyearpar{ThompsonONTL} with Beta priors applied to bandits with expert advice is $\Theta(\sqrt{TNlogT} + \zeta T)$, where the first term is by \cite{lattimore2020bandit}, and the second term accounts for the best expert's gap to optimality.

\subsection{EXP4-IX}
EXP4-IX incurs a regret upper bounded by $ O(\sqrt{KTlogN} + \zeta T) $ \citep[adjusted again for the best expert's gap to optimality]{neu2015explore}. 
Hence, while the MAB reduction had no dependence on the number of arms $K$, this dependence is now introduced in the EXP4-IX algorithm. The upside however is a lower dependence on both the number of time steps and the number of experts. 

\subsection{Meta-CMAB}

We consider several CMAB algorithms and their applicability to the case wherein contexts are expert advice, i.e., estimates of the true outcomes. While regret bounds are known for the CMAB algorithms we consider, we study here their validity for the case wherein the context consists of expert advice. 

\subsubsection{ILTCB}
As analyzed by \cite{agarwal2014taming}, ILTCB, when applied on the expert set, achieves a regret of $$ O(\sqrt{KT log(T N/\delta)} + K log(TN/\delta) + \zeta T) $$

While this bound is worse than EXP4, ILTCB is more efficient, and thus might be more practical when applied to larger policy classes. ILTCB however does assume a finite policy class, and bounds therefore would not hold for the linear policy class. A finite approximation of the linear policy class in the form of a cover results in a policy class whose size is exponential in the number of experts, and thus has regret $$O(\sqrt{KT Nlog(T/\delta)} + KN log(T/\delta) + \varepsilon T )$$

This results in an undesirable linear dependence on both the number of arms and experts.

\subsubsection{SafeFALCON}

Central to SafeFALCON's analysis is the existence of an offline oracle with appropriate learning guarantees. In particular, \cite{Krishnamurthy2021} establish a regret of $O(\sqrt{K\xi(T,\delta/log(T))}T + \varepsilon \sqrt{K}T)$, wherein $\xi(T,\delta/log(T))$ is the oracle's estimation rate, i.e., an upper bound on its excess square error which holds with probability $1- \delta/log(T)$ when fit on $T$ experiences. 

For example, for linear least squares, this quantity can be shown to be $O((d-log(\delta/logT))/T)$ under very weak assumptions \citep{audibert2009risk,koltchinskii2011oracle}. By substituting $d$ for the actual meta-context dimensionality $N$, and setting $\delta=logT/T$, the resulting regret is $$O(\sqrt{KT(N+logT)} + \varepsilon \sqrt{K}T)$$
This significantly improves on ILTCB applied to a finite approximation of the linear space. 




\subsubsection{Sup-Lin-UCB-Var}

In order to provide a bound on the regret of Sup-Lin-UCB-Var applied to the Meta-CMAB, some quantities need to be bounded as follows. Each meta-context, i.e., the $N$-sized vector of advice for each arm is bounded as $||\vv{\tilde{f}}||_2 \leq \sqrt{N}$. In particular, this norm is maximized if all experts agree that a given arm's value is $1$. We further make a relatively weak assumption that the optimal weights are bounded as 
$||\theta^* ||_2\leq 1$. Which is for example the case for any optimal weighted average. We also have that rewards are in $[0,1]$, and therefore the maximal gap in terms of arm outcomes is also bounded: $\max_{k,k'\in[K],t\in [T]} | \mathcal{E}(\vv{\tilde{f}}_{k,t}) -\mathcal{E}(\vv{\tilde{f}}_{k',t})| \leq 1 $.
Finally, the analysis of Sup-Lin-UCB-Var assumes conditionally sub-gaussian noise. As discussed previously, this can be shown to be equivalent to noisy expertise (i.e., features), assuming the noise on expertise is itself conditionally sub-gaussian.
With these bounds and assumptions in place, it is straightforward to apply the analysis of \citep{takemura2021parameter} to the Meta-CMAB with dimensionality $d=N$, which leads to a regret of 
$$\tilde{O}(\sqrt{NTlog(K)} + \varepsilon \sqrt{N}T)$$ wherein $\tilde{O}(\cdot)$ ignores polylogarithmic factors in $N$ and $T$. 
Compared to SafeFALCON, this bound has a better dependence on the number of arms. We remark however that the penalty of misspecification grows with $\sqrt{N}$, rather than with $\sqrt{K}$ for SafeFALCON.

\ 

We have so far expressed regret bounds in function of either $\varepsilon$ or $\zeta$.
Given these two measures of misspecification, optimizing over the larger linear class is only justified when the gap between $\varepsilon$ and $\zeta$ is sufficiently large, i.e., when the optimal linear aggregation sufficiently outperforms the single best expert. In order to gain an intuition as to when this is the case, we establish some results bounding the potential improvement in what follows.

 \subsection{Estimating Misspecification}\label{sec:misspecification}
 While the best linear combination of expert advice is necessarily at least as good as the single best expert, some misspecification is likely to exist. In what follows we establish some bounds on the level of misspecification for particular cases of expertise. 
 While an unconstrained linear regression is more flexible than a weighted average, we note that i) negative experts can be accommodated by including an inverted copy of each expert, and ii) \cite{breiman1996stacked} found that typically, when features are the estimates of sub-regressors, the sum of weights tends to converge towards $1$. This is in part enforced by regularisation, which, in particular, avoids excessive over-fitting when experts are highly correlated. The results we establish here by assuming $\vv{\theta} \in \Delta([N])$ thus provide a tight upper bound on the best unconstrained linear model's error.
 
We start by restating the error of linear models. 
 Let $\mathbf{C}$ be the covariance matrix of residuals, such that 
 $$C_{nm} = \frac{1}{KT}\sum_{t=1}^T\sum_{k=1}^K  (f(k,\vx_t) -\tilde{f}^n(k,\vx_t))(f(k,\vx_t) -\tilde{f}^m(k,\vx_t))$$ and let $\vv{\theta}\in \Delta([N])$ be a weight vector, the error of the linear aggregation is then \citep{breiman1996stacked}:
 $$ \epsilon(\vv{\theta}) = \frac{1}{KT}\sum_{t=1}^T\sum_{k=1}^K  (f(k,\vx_t) - \langle \vv{\theta} , \vv{\tilde{f}}_{k,t} \rangle )^2  = \sum_{n=1}^N \sum_{m=1}^N \theta_n \theta_m C_{nm}$$
 We will omit the parameter of $\epsilon$ when it is clear from the context. 
 This general result indicates that model error decreases as the covariance of experts ($C_{nm}\; \forall n\neq m$) decreases, and as the error of experts ($C_{nn}$) decreases. Therefore, for a fixed level of expertise, an increase in diversity (as measured by a decreased covariance), improves the performance obtainable through the collective, and thus improves the performance of our Meta-CMAB approach.
 
 If the covariance matrix is invertible, the optimal weights can be obtained by 
 $$ \theta^*_n = \frac{\sum_{m=1}^N (C^{-1})_{nm}}{\sum_{o=1}^N \sum_{m=1}^N (C^{-1})_{om}} $$

 We will denote by $\chi(\mathbf{C})= \min_{o\in [N]} C_{oo} - \epsilon(\vv{\theta}^*) $ the linear model's improvement over the single best expert.
We can relate this value back to the misspecification defined in \autoref{sec:theoretical}, specifically $\varepsilon$ for the linear model's misspecification and $\zeta$ the expert misspecification. 
Given a covariance matrix $\mathbf{C}$ and the linear model's improvement $\chi(\mathbf{C})$, we have that 
$$ \mathds{E}[\varepsilon - \zeta] =\sqrt{\chi(\mathbf{C}) } $$

 From these general results we can derive some special cases which are likely to occur for example when experts are human.
 
 \subsubsection{Special Cases}
 
 \paragraph{Independent Experts}
 If experts are independent, such that $C_{nm}=0\; \forall n\neq m$, the optimal weight for expert $n$ is \citep{sinha2011statistical}:
 $$\theta^*_n = \frac{1/C_{ii}}{\sum_{m=1}^N 1/C_{mm}}$$
 
 And the model error is

 $$\epsilon = \frac{1}{\sum_{n=1}^N 1/C_{nn}} $$ 
 
The linear model's improvement over the single best expert is then
 $$\chi(\mathbf{C}) =  \frac{\sum_{n\neq o} C_{oo}/C_{nn}}{\sum_n 1/C_{nn}}  $$
 Which grows as the error of the best expert ($C_{oo}$) increases. Conversely, as $C_{oo}\rightarrow 0$, the usefulness of an aggregate declines.

 \paragraph{Homogeneous Experts}
 Assume all experts display similar expertise, and correlation across experts is homogeneous.  This configuration can capture for example trained professionals whom one might expect to be relatively similar in terms of expertise. Specifically, let $C_{nn} = a \;\forall n$ and $C_{nm}= c \;\forall n \neq m$. The simple average is then optimal (i.e., $\theta^*_n = 1/N \;\forall n$), and the expected error is $$ \epsilon =  \frac{a - c}{N} + c $$
 We therefore pay a constant price ($c$) due to correlation, which does not decrease as we increase the number of experts. In contrast, the loss induced by individual errors ($a$), decreases as we increase the number of experts. 
 If all experts are independent ($c=0$), the model error is $a/N$. As we increase the number of experts, we reduce the model error, which is however offset by the increased complexity induced by the higher number of experts. 
 Conversely, if experts are essentially identical (i.e., $c \approx a$) the expert set is de facto equivalent to a single expert, and the resulting model error is $\epsilon = a$.
 
 We can again quantify the improvement of an aggregate over the single best expert as
 $$\chi(\mathbf{C}) = a - \big(\frac{a - c}{N} + c\big) = (a - c)\frac{N-1}{N} $$ 
 Which increases as the number of experts increases, converging towards $a - c$. Therefore an aggregate is most beneficial when the number of experts is large, the error of individual experts is large, and the correlation between experts is low.

 \paragraph{Heterogeneous Experts}
 Experts from different backgrounds who are not carefully chosen can vary widely in terms of performance. Applications relying on crowd-sourcing are prime examples of such heterogeneity.  
 Consider the case wherein expertise is uniformly distributed, such that $C_{ii} = a+(i-1)(1-a)/(N-1) \;\forall\, i\in [N] $. 
 Assuming experts are independent ($C_{ij}=0\; \forall\, i\neq j$), we can derive the model's misspecification as being $$\chi(\mathbf{C}) = C_{11}   -\frac{1}{\sum_{n=1}^N 1/C_{nn}}   = a - \frac{a-1}{N H_N }  $$
 wherein $H_N$ is the $N$th harmonic number. Therefore, in the heterogeneous case, we again have that the linear model's improvement grows as the number of experts increases, or as the smallest error ($a=C_{11}=min_{o\in [N]}C_{oo}$) increases.

 \paragraph{Polarized Experts}
 \emph{Polarized} collectives are marked by disagreeing clusters of similar experts. Such clusters can for example occur in policy-making along the left-right axis \citep{poole1984polarization}. These polarized experts are characterized by high inter-cluster correlation, and low intra-cluster correlation.
 Assume experts are divided into $L$ disjoint clusters, i.e., sets of indices such that for every cluster $ \mathcal{L}\in L$, $C_{ij}=c_\mathcal{L} \;\forall\, i\neq j \in \mathcal{L}$ and $C_{ii}=\epsilon_\mathcal{L} \;\forall\, i \in \mathcal{L}$. And assume these clusters are independent s.t., $C_{ij} = 0 \;\forall\, i\in \mathcal{L}, j\in \mathcal{L}'\neq \mathcal{L}$. 
 
Because these clusters are independent, the optimal aggregation is a weighted average of the average of clusters. Specifically, following previous results, we can establish the expected error of each cluster $\mathcal{L}$ as being:
$$ a_\mathcal{L} = \frac{\epsilon_\mathcal{L} - c_\mathcal{L}}{|\mathcal{L}|} + c_\mathcal{L}$$

The optimal weights for a given cluster is then:

$$ \theta_\mathcal{L}^* = \frac{a_\mathcal{L}^{-1}}{\sum_{\mathcal{L}'}a_\mathcal{L}^{-1}} $$

And the expected error is:
\begin{equation} 
\begin{split}
\sum_{\mathcal{L}} (\theta^*_{\mathcal{L}})^2 a_\mathcal{L} & = 
    \sum_{\mathcal{L}} \big(\frac{a_\mathcal{L}^{-1}}{\sum_{\mathcal{L}'}a_\mathcal{L'}^{-1}}\big)^2 a_\mathcal{L} \\ 
    & =     \frac{\sum_{\mathcal{L}}a_\mathcal{L}^{-1}}{\big(\sum_{\mathcal{L}'}a_\mathcal{L'}^{-1}\big)^2} \\ 
     & =  \frac{1}{\sum_{\mathcal{L}}a_\mathcal{L}^{-1}} \\ 
     & =  \big(\sum_{\mathcal{L}}(\frac{\epsilon_\mathcal{L} - \sigma_\mathcal{L}}{|\mathcal{L}|} + \sigma_\mathcal{L})^{-1}\big)^{-1}
     \end{split}
\end{equation}
 
 This value indicates that --- all other factors being equal--- the error tends to decrease when i) the size of clusters increases, ii) the number of clusters increases, or iii) (co)variance within clusters decreases.
 
 Note that the homogeneous case discussed previously is equivalent to a single cluster of size $N$. In contrast, the heterogeneous configuration is equivalent to $N$ clusters of size $1$.

 \subsubsection{From model error to misspecification}
The expected errors we derived so far differ from misspecification, which corresponds to the expected maximum error, but can be used to bound misspecification.

 \paragraph{Adversarial error} Without making any assumptions, the worst-case misspecification occurs when the error is concentrated on a single arm. Given an expected (squared) error of $\epsilon$ over $K$ arms, the worst case misspecification is  bounded as $\varepsilon=O(\sqrt{\epsilon K})$. 
 
 \paragraph{Stochastic error}
 In the less general setting wherein errors are i.i.d. $\sqrt{\epsilon}$-sub-gaussian, the misspecification is the expected maximum error over $K$ arms, which is upper bounded by $ \sqrt{\epsilon 2 log2K}=O(\sqrt{\epsilon logK})$ \citep[Lemma 2.3]{massart2007concentration}.


 
\section{Experimental setting}
The aim of this experimental setting is to evaluate the performance of Meta-CMAB compared to state-of-the-art algorithms for a wide variety of configurations involving either synthetic or crowd-sourced experts (derived from \citet{whitehill2009whose,buckley2010overview,snow2008cheap})\footnote{Code to reproduce these results is available at \url{https://github.com/axelabels/CDM_BIAS}.}. Specifically, we wish to empirically confirm the effectiveness of using linear aggregations over the set of experts and evaluate the effectiveness of the different exploration strategies employed by the algorithms presented in \autoref{sec:metacmab}.
Because of the lack of publicly available data sets which would allow us to extensively experiment with biases, number of experts, and number of arms, we first  adopt a synthetic experimental setting. Following this, we simulate instances involving human experts by adapting crowd-source data sets to the setting of bandits with expert advice. We report performance in terms of average reward. CDM algorithms must thus rapidly learn to use the (possibly biased) advice provided by experts. 
We explore varying expert configurations detailed in \autoref{sec:expert_configs}.
Given our concern with providing effective algorithms for real-world decision problems, we focus on settings involving relatively few time steps ($T\in \{10,100,1000,10000\}$), experts ($N\in \{4,8,16,32,64,128\}$) and arm counts ($K\in \{4,8,16,32,64,128\}$).

\subsection{Synthetic Bandits}
We consider two bandit types which mirror the classic supervised learning problems of regression and classification.
\paragraph{Regression}
Regression bandits are characterized by a reward function which is not concentrated on a single arm. Specifically, we consider bandits wherein the expected reward for arms is distributed uniformly on $[0,1]$. 

\paragraph{Classification}

In addition to regression bandits, we consider bandits wherein the difference in reward between the single best arm and all other arms is much more pronounced, i.e. classification bandits. For any given time-step a single arm (the one corresponding to the appropriate class) has reward $1$, and the other arms have reward $0$. This is a significant change from the regression bandit in a few ways. First, the expected reward for a policy choosing arms at random is $1/K$, as opposed to $1/2$ in the case of regression bandits. And secondly, this changes how adversity affects performance. In the absence of correlation between experts, erroneous advice is likely to be spread out amongst the $K-1$ pessimal arms, while accurate advice will coincide with the single best arm. It is therefore this single best arm that will garner most support and be chosen. This implies that a weaker plurality of accurate experts can outvote a majority of uncorrelated and error-prone experts.

\subsubsection{Biased Expertise}\label{sec:expertise_bias}
We generate biased expertise as follows.
For regression bandits, let $\mathbf{\eta}^n \sim \mathcal{U}(0,1)^{t\times K}$ be expert $n$'s noise sampled from the uniform distribution. Let $f_{k,t}$ be the expected outcome of arm $k$ at time $t$, and let $\epsilon_n$ be the error of expert $n$. 
To generate erroneous advice, we let expert $n$'s approximation of arm $k$'s expected reward at time $t$ be $\tilde{f}^n_{k,t} := f_{k,t}w_1(\epsilon_n) +  (1-f_{k,t}) w_2(\epsilon_n) + \eta_{k,t}^n w_3(\epsilon_n)$. With $w_1(\epsilon) := max(0,1-2\epsilon)$, $w_2(\epsilon) := max(0,2\epsilon-1)$, and $w_3(\epsilon) := 1 - w_1(\epsilon) - w_2(\epsilon)$. Thus, an expert's advice is a mixture of the (possibly inverted) true outcome and some noise term. Specifically, at $\epsilon=0$, the advice is identical to the expected outcome, at $\epsilon=0.5$, the advice is sampled randomly from a uniform distribution, and at $\epsilon=1$, the advice is the complement of the expected outcome. 
For classification bandits, we can expect an expert's advice to sum to $1$ (as a single arm is correct). For each time step, we let each expert provide the correct advice with probability $1-\epsilon_n$, and otherwise sample a random valid (i.e., summing to $1$) advice vector.

With this measure of distance, we also impose a correlation between experts, elaborated in the following section.

\subsubsection{Expert Configurations}\label{sec:expert_configs}
We consider configurations which mirror those analyzed in \autoref{sec:misspecification}.
First, situations wherein experts are globally \emph{homogeneous} in their expertise, i.e., they are all approximately equally close or far from the truth. In contrast, \emph{heterogeneous} experts can vary significantly in their expertise. 
In addition to these two configurations, we also consider how correlation affects performance in the case of polarized collectives.

Let $\epsilon$ be a desired level of expertise which characterizes the performance of the set of experts. Individual expertise, as defined by a value $\epsilon_n$ for each expert $n$, differs according to the expert configuration as follows.

\paragraph{Homogeneous experts}
Homogeneous experts have similar levels of expertise. That is, $\epsilon \approx \epsilon_1 \approx \epsilon_2 \approx ... \approx  \epsilon_N $

\paragraph{Heterogeneous experts}
Heterogeneous experts are uniformly spread across the spectrum of expertise. Specifically, $\epsilon_i \approx 0.2+(i-1) 0.8\epsilon/(N-1)$.

\paragraph{Polarization}
Polarization levels are defined by a coefficient $p$ which determines how strongly experts within a cluster are correlated. Specifically, let $\rho_{ij}$ be Pearson's correlation coefficient between the advice of experts $i$ and $j$. We simulate polarization by sampling experts such that 
\[
    \rho_{ij} = 
\begin{cases}
    0,& \text{if } \mathds{1}_{i\leq cN} \neq \mathds{1}_{j\leq cN} \\
   p ,              & \text{otherwise}
\end{cases}
\]
Where $c$ is the proportionality of the first cluster's size (such that $cN+(1-c)N = N$).

Experts are thus subdivided into two clusters within which experts share a correlation coefficient of $p$.
For example, in the $4$ experts case, and for $c=0.5$, the correlation matrix is given by
 $$\begin{bmatrix}
1 & p & 0 & 0 \\
p & 1 & 0 & 0 \\
0 & 0 & 1 & p \\
0 & 0 & p & 1 \\
\end{bmatrix}  $$
 High values of $p$ induce high polarization, i.e., strongly correlated subsets of experts.

Given the desired correlation level $p$ of a cluster of experts, and $\epsilon$ their expected error, we achieve correlation by sampling advice from a common distribution of $\epsilon$-biased advice with probability $p$. 

Specifically, let $\mathbf{\tilde{f}}^{i}\in [0,1]^{T \times K}$ be a matrix of shared beliefs, i.e., advice common to the $i$-th cluster. The polarized advice provided by expert $n \in C_i$ at time $t$ for arm $k$ is $\tilde{f}_{k,t}^{n} =(1-B_{k,t}^{n})\tilde{f}_{k,t}^{n} + B_{k,t}^{n}(\tilde{f}_{k,t}^{i})$ with $B_{k,t}^{n}\sim {Bernoulli}(p)$. In other words, there is a probability of $p$ that an expert follows the shared beliefs rather than its own when providing advice.

\subsection{Human Expertise Data Sets}
In addition to the synthetic setting, we will also validate our results on problems involving human expertise. To do this we evaluated on three data sets consisting of classification or regression problems solved by crowd-sourced experts. 

\paragraph{Classification data sets} In a classification task, humans are repeatedly presented with a problem instance (i.e., a context), and asked to assign an appropriate label to the problem, for example whether an excerpt is relevant to a given topic \citep{buckley2010overview}. We transform such classification tasks to bandits in the same manner we transformed the above data sets. Specifically, each possible label is an arm, and we collect a reward of $1$ if we choose the true label of the given instance. As human experts were asked to choose a single label, the advice of an expert at each round is concentrated on a single arm. The first classification task on which we evaluate --- based on the \emph{Duchenne data set} \citep{whitehill2009whose} --- asks human experts to evaluate whether a smile is real or fake (hence $2$ labels/arms). The second, the \emph{Recommendation} data set \citep{buckley2010overview}, asks whether a web page is relevant to a given topic, with possible answers being highly relevant, relevant, or non-relevant ($3$ labels/arms). 

\paragraph{Regression data set} The \emph{Emotion data set} \citep{snow2008cheap} asks human experts to rate the presence of different emotions (anger, disgust, fear, joy, sadness, surprise) in a picture on a $0-100$ scale. We transform this data set into a $6$-armed bandit (one arm per emotion), wherein the aim is to select the strongest emotion. An expert's advice in this case consists of the provided rating for each emotion (i.e., arm). 

\paragraph{Data set use}
From the set of available experts, we select for each trial a random subset of $4$ experts and each round consists of one of the problem instances for which all $4$ experts provided an answer. For the Duchenne and recommendation data sets, ground truths are often absent. To handle this absence we sample a $5$th expert whose advice is substituted for the ground truth. The number of rounds is limited to the number of common problem instances solved by the chosen group of experts. For the Recommendation data set this results in approximately $30$ rounds per set of experts, approximately $250$ rounds per set of experts for the Duchenne data set, and $20$ rounds per expert set for the emotion data set. For each expert set, we run our algorithms on $10$ random permutations of the available problem instances.

While these data sets do not give us the opportunity to extensively test for different expert configurations, the diversity of the available experts can be evaluated in terms of their standard deviation. For smaller standard deviations we obtain homogeneous configurations, while larger standard deviations are similar to heterogeneous configurations.

\section{Results and Discussion}

We evaluate performance in terms of average reward.
In addition to the performance of different algorithms, we also plot the performance of the best and worst expert. Algorithms we evaluate are the following. Meta-MAB, which reduces the task to a simple multi-armed bandit and solves it using Thompson Sampling. EXP4-IX applied to the set of greedy policies induced by the experts, i.e., each policy acts greedily on its corresponding expert's advice. The Average method greedily acts on the average of expert advice. In addition to these baselines, we compare different approaches to solving the Meta-CMAB as discussed previously, using Safe-FALCON (Meta-CMAB (FALCON)), using Online Cover, the online approximation of ILTCB (Meta-CMAB (ILTCB)), and finally using the variant of LinUCB robust to misspecification (Meta-CMAB (UCB)).

\subsection{Classification compared to regression}

\begin{figure}
\centering
 
 \includegraphics[width=1\textwidth]{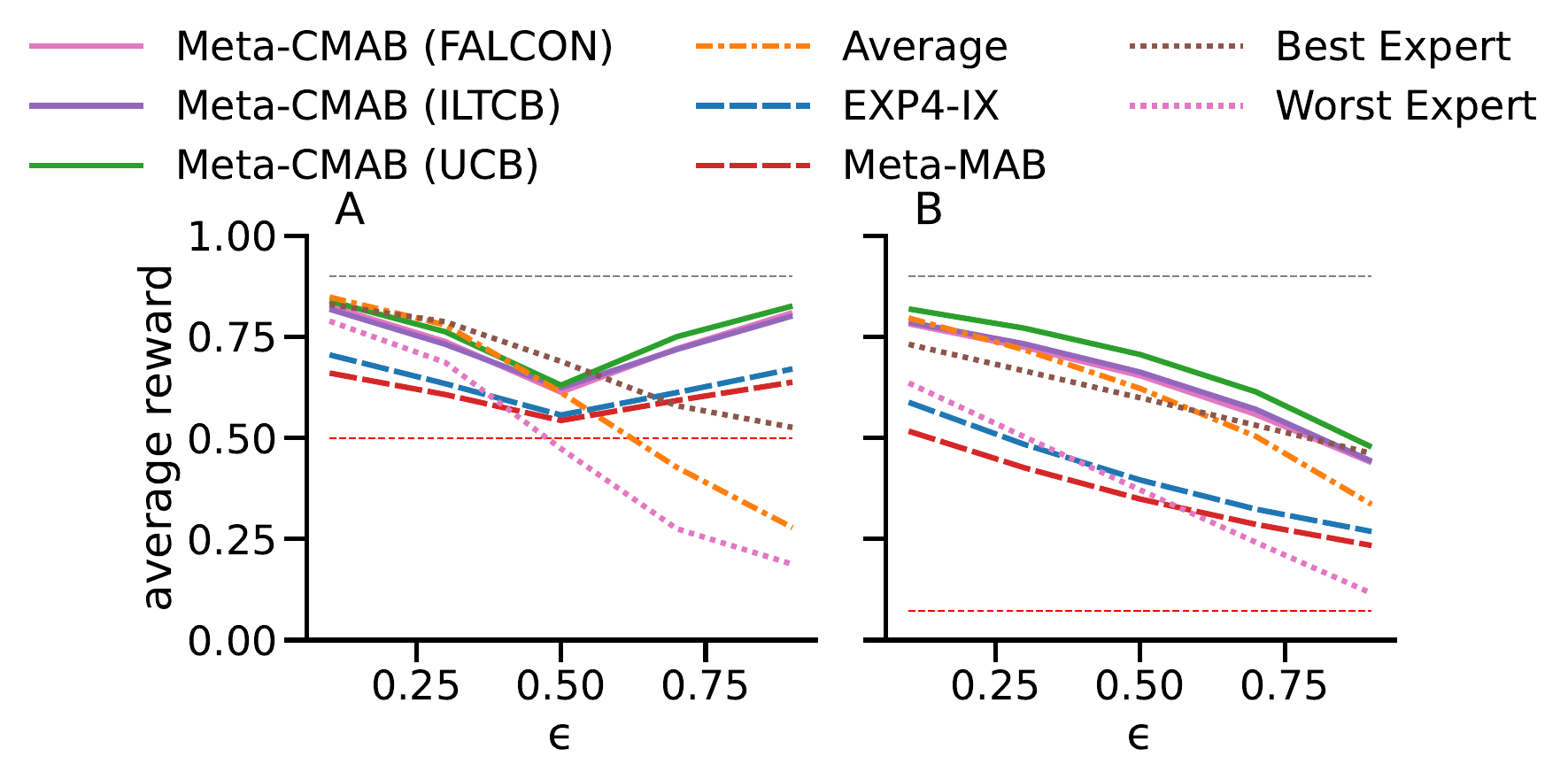}

 \caption{%
 Meta-CMAB compared to state-of-the-art on A) \textbf{regression} or B) \textbf{classification} bandits. Average reward in function of expert misspecification (i.e., $\epsilon$). \textit{Best/Worst expert} is the performance obtained by the best/worst expert of each run. %
 The grey and red dashed lines mark the expected performance of respectively an optimal and random policy. %
 }\label{fig:type_comparison}
\end{figure}
We first provide an overall comparison of the performance of our different algorithms on classification or regression bandits. Results presented in \autoref{fig:type_comparison} thus average performance over all possible configurations (heterogeneous or homogeneous), all polarization levels, and all combinations of $N$ and $K$. The general conclusions we draw from this comparison will also hold for more detailed results presented in the subsequent sections. 

We find that overall, \emph{Meta-CMAB (UCB)} maximizes performance among all aggregation algorithms. In particular, it significantly outperforms EXP4-IX, Meta-MAB and the simple average in the classification case, and in the regression case when worse-than-random experts are present. When all experts are better-than-random ($\epsilon<0.5$), the simple average provides competitive performance. It is interesting to note that in the latter case, the simple average is close to optimal, yet Meta-CMAB (UCB) is able to approach its performance in relatively few steps. Remark that, if we have certainty about the presence of only better-than-random experts, we can induce appropriate priors on the linear weights learned by our Meta-CMAB methods. This would ensure that, initially, Meta-CMAB performs as a simple average would, and then converges towards a weighted average to deliver improved performance over the simple average. 

In terms of Meta-CMAB variants, we find that the UCB variant consistently outperforms the other variants, followed by the FALCON and ILTCB variants. This suggests that specialized exploration driven by variance estimates (as used by the UCB approach) is effective in this setting. 

Finally, comparing two more traditional approaches for deciding with expert advice (namely EXP4-IX and Meta-MAB), we find they provide performance halfway between that of a random policy and that of \emph{Meta-CMAB (UCB)}. As these two algorithms act on a single expert's advice, they are unable to leverage the collective intelligence the Meta-CMAB approaches can. The seemingly better-than-best performance obtained by these algorithms in some case is a consequence of the inversion method. That is, for each expert, a synthetic expert whose advice is inverted is included. While this allows these methods to surpass the best expert in some cases by exploiting the best inverted expert instead, it also induces reduced performance when experts are better-than-random. In this latter case, inverted experts are worse-than-random and thus only slow down EXP4-IX and Meta-MAB's convergence towards the single best expert. 

The reward distribution induced by the classification problem results in a lower threshold to perform better than random (as highlighted by the dashed red line in \autoref{fig:type_comparison}). As a consequence, our Meta-CMAB approach can efficiently exploit the collective even when a minority of experts are accurate for any given round. In such uncorrelated configurations, inaccuracies tend to be distributed evenly across sub-optimal arms, and a simple plurality of accurate advice is sufficient to put decisive mass on the appropriate arm. In contrast, Meta-MAB and EXP4-IX cannot exploit such pluralities, but attempt instead to identify the single best expert, which leads to worse performance.

\subsection{How does the distribution of expertise influence performance?} \label{sec:results_distance}

\begin{figure}
\centering
 
 \includegraphics[width=.9\textwidth]{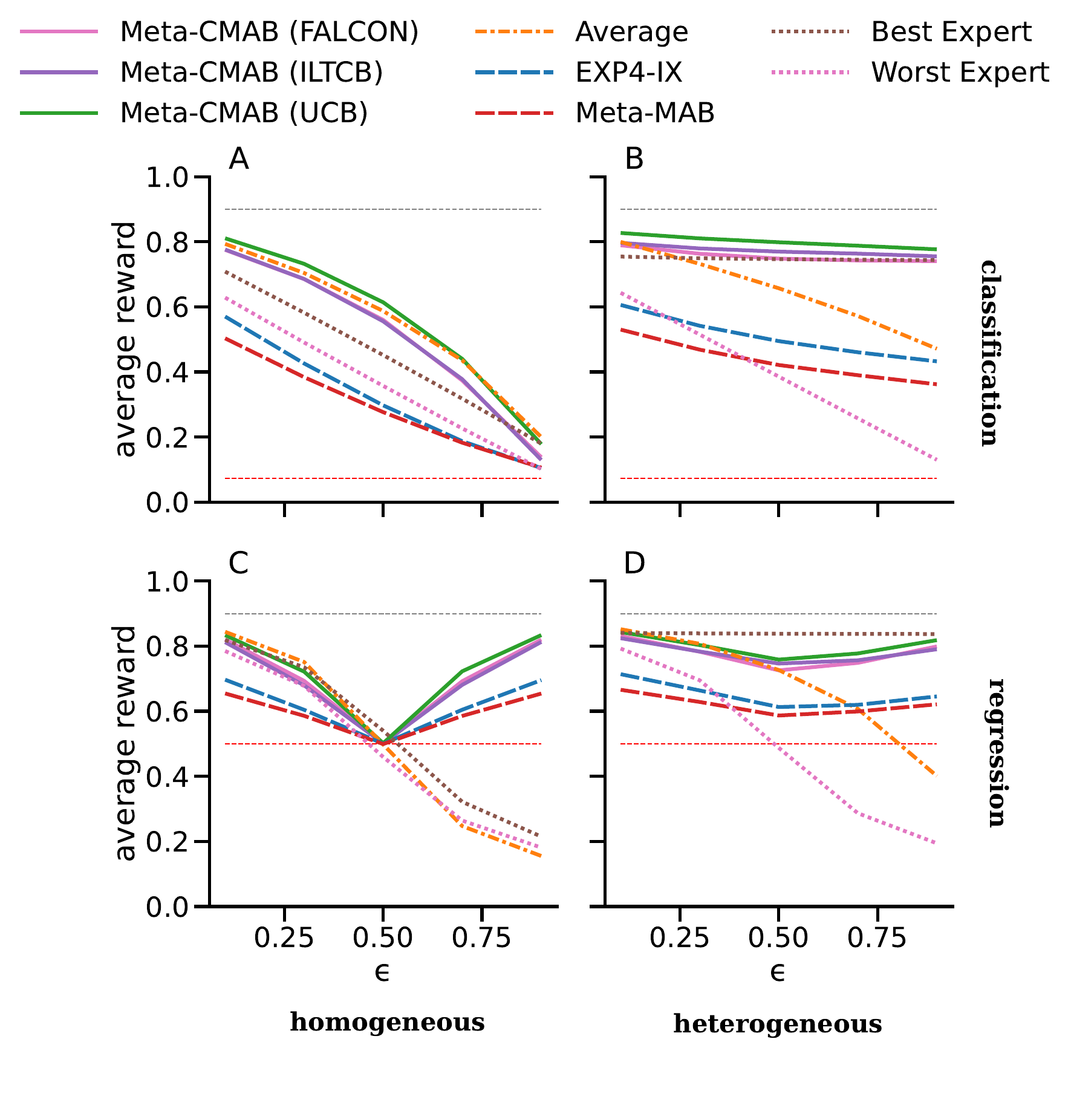}

 \caption{%
 Meta-CMAB compared to state-of-the-art on (top) \textbf{classification} or (bottom) \textbf{regression} bandits, and for either (left) \textbf{heterogeneous} or (right) \textbf{homogeneous} expertise. Average reward in function of expert misspecification (i.e., $\epsilon$). \textit{Best/Worst expert} is the performance obtained by the best/worst expert of each run. %
 The grey and red dashed lines mark the expected performance of respectively an optimal and random policy. %
 }\label{fig:type_distr_comparison}
\end{figure}

As we laid out in \autoref{sec:expert_configs}, different expert configurations correspond to different distributions of errors in the experts. %

\begin{figure}
\centering

 \includegraphics[width=.49\textwidth]{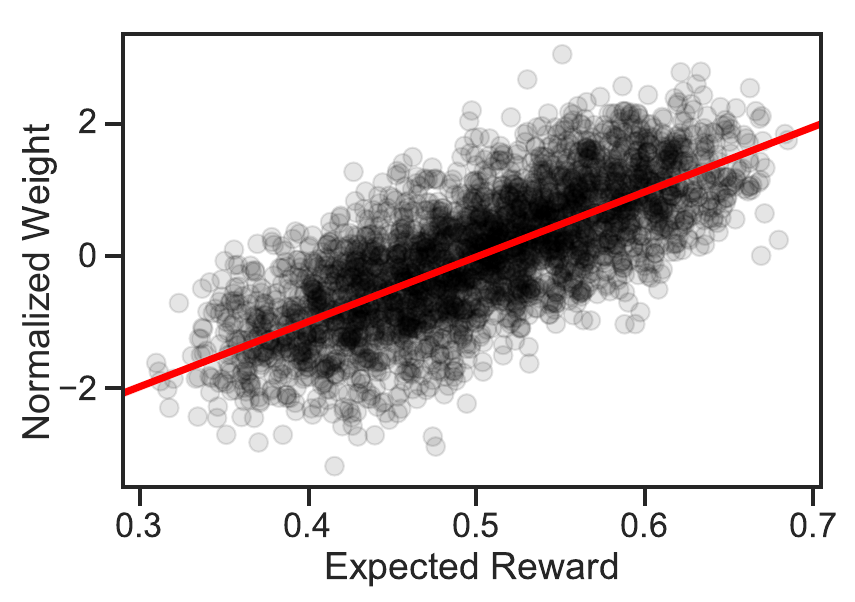}
 \includegraphics[width=.49\textwidth]{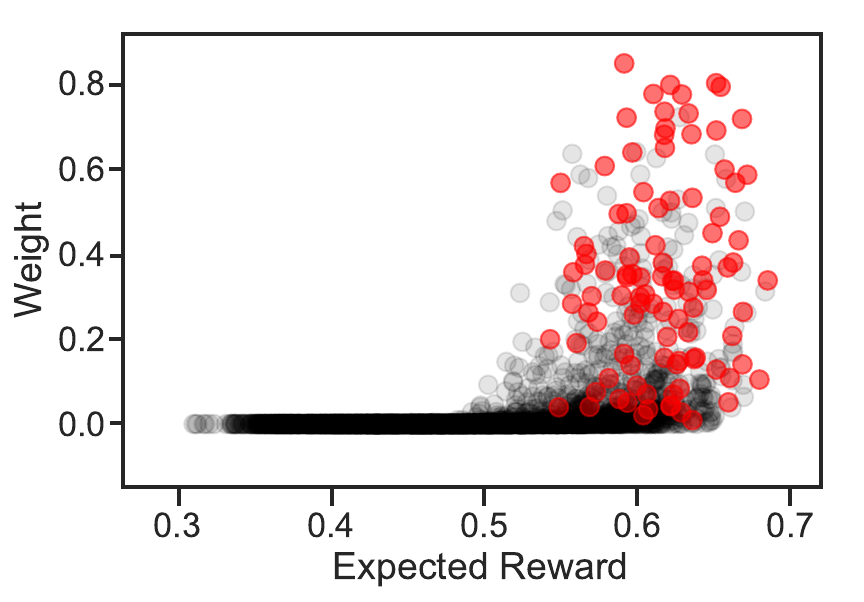}
 
 \caption{Expert importance in function of their expected reward. For 100 runs with sets of 32 experts, each expert's expected reward is computed in hindsight and plotted against the expert's weight in the model. \textbf{(left)} In the Meta-CMAB case the weights are the normalized weights in the linear model. A linear regression (red) is fitted to the Meta-CMAB data. \textbf{(right)} For EXP4-IX the weights are the exponential weights scaled such that the sum of weights for the set of experts is $1$. The single best expert for each run is highlighted in red. }\label{fig:weights_pcc}
\end{figure}

\autoref{fig:type_distr_comparison} shows how these configurations affect the average reward for the considered algorithms for either regression or classification bandits.
In the homogeneous case, we observe a decline in performance of both the state-of-the-art and Meta-CMAB algorithms as the expertise declines. Interestingly, as the difference goes beyond $0.5$ one can observe that algorithms recover some of their performance in the regression case. To explain this, we remark that the performance of a random policy on regression bandits is approximately $0.5$, hence it is at this point that expertise is least correlated with the collective bandit. For $\epsilon$ beyond $0.5$, expertise becomes negatively correlated. This negative correlation is identified and exploited by the different algorithms with varying levels of success. 
In particular, the Meta-CMAB approach learns a linear relation between advice and expected outcome. When expertise is weakly correlated, this relation is harder to establish. In contrast, when expertise is strongly negatively correlated, the inverse relation between advice and actual outcome is identified and exploited by Meta-CMAB. The left-hand plot in \autoref{fig:weights_pcc} supports the hypothesis that there is a correlation between an expert's expected performance and the expert's weight in the linear model learned through Meta-CMAB. Concretely this implies that the advice of experts with a negative correlation is inverted, and that the experts with a higher absolute correlation have a higher importance. In practice, we find that this allows \emph{Meta-CMAB} to perform strongly in all but one case. In particular, only when all experts are completely random (\autoref{fig:type_distr_comparison}.C, $\epsilon=0.5$) does the Meta-CMAB fail to learn how to act on expert advice. This however is inevitable, as there is no correlation between advice and outcomes. The advantages of Meta-CMAB are particularly strong for the classification case. This is in part because a plurality is sufficient, but also results from the lower variance in the reward signal (induced by the more eccentric values $f$ takes in the classification bandit). 

 When simply using the provided expert advice, adaptive algorithms like Meta-MAB and EXP4-IX would track the performance of the best expert. Yet, by including inverted copies of all experts, a similar exploitation of weak experts is achieved by these adaptive methods. This results in the same upward trend past $\epsilon=0.5$. As we previously discussed, in exponential weights methods, such as EXP4-IX, the weights tend to concentrate on a single expert. The right-hand plot in \autoref{fig:weights_pcc} illustrates this property, as negatively correlated experts are assigned a weight close to $0$. While EXP4-IX does tend to converge towards good experts, its policy consists in acting on a single expert's advice. In practice this prevents collective intelligence and as shown here, results in reduced performance when compared to Meta-CMAB. 
The reduction to a multi-armed bandit, Meta-MAB, is similarly limited to the use of a single expert, and the resulting impact on performance are confirmed here. Finally, as can be expected from the literature, acting on the average of expert advice can produce collective intelligence when experts are better than random and (as is necessarily the case for the homogeneous experts) similar in performance. 

In contrast with the homogeneous configuration, the heterogeneous configuration offers the CDM algorithms a wider variety of experts in terms of their prior bandit's distance to the collective bandit. In practice this results in the availability of better experts than for the homogeneous configuration. Learning algorithms (EXP4-IX, Meta-MAB and Meta-CMAB) are able to identify and use these experts in their decision making, which results in improved performance, displayed in \autoref{fig:type_distr_comparison}. The different algorithms are able to exploit the better experts to varying degrees, resulting in improved performance. Furthermore, the various algorithms benefit not only from strong experts, but also from the worse-than-random experts, as both are strongly (positively or negatively) correlated to the collective bandit. 

\subsection{Polarization induces a performance loss}

\begin{figure}
\centering
 
 \includegraphics[width=1\textwidth]{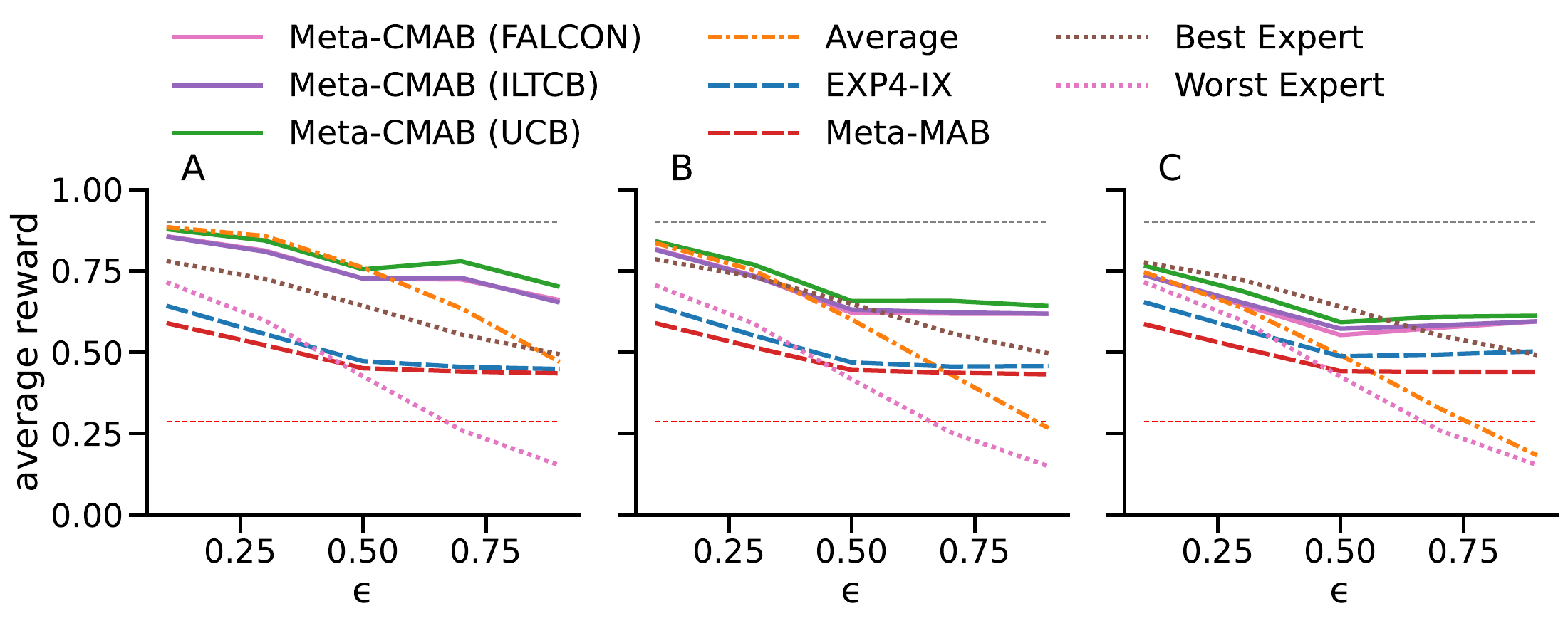}

 \caption{%
 \textbf{Homogeneous} experts, Meta-CMAB compared to state-of-the-art on \textbf{classification} bandits. Average reward in function of expert misspecification (i.e., $\epsilon$) for different levels of \textbf{polarization}; A) low polarization ($\rho=0.1)$), B) medium polarization ($\rho=0.5)$), and C) high polarization ($\rho=0.9)$). \textit{Best/Worst expert} is the performance obtained by the best/worst expert of each run. %
 The grey and red dashed lines mark the expected performance of respectively an optimal and random policy. %
 }\label{fig:correlation_error}
\end{figure}

As the results we discuss in this section are similar across configurations, we present here the results averaged across configurations in \autoref{fig:correlation_error} and defer to supplementary figures \ref{fig:class_correlation_error_hom} to \ref{fig:regr_correlation_error_het} for the detail per configuration. 

A trend shared by the Meta-CMAB variants and the Average in these results is a decrease in performance as the polarization increases. Higher polarization induces a higher correlation between experts of the same cluster. This correlation produces redundancy within the experts, which in turn results in a weaker collective. For high correlation, the expert set is essentially reduced to two clusters of almost identical experts. The decreased performance thus follows from the larger model error induced by high polarization (as discussed in \autoref{sec:misspecification}). 

In contrast, polarization does not have a significant negative impact on EXP4-IX and Meta-MAB. As these algorithms do not exploit the collective, but rather converge towards a single expert, they are simply bounded by the performance of the best (inverted) expert. Thus, Meta-MAB appears constant in its performance. Note however that EXP4-IX slightly improves in performance, as the agreeing experts facilitate convergence.

\subsection{Effect of changes in the number of arms and experts} 

\begin{figure}

     \begin{subfigure}[b]{0.48\textwidth}
\centering
 
 \includegraphics[width=1\textwidth]{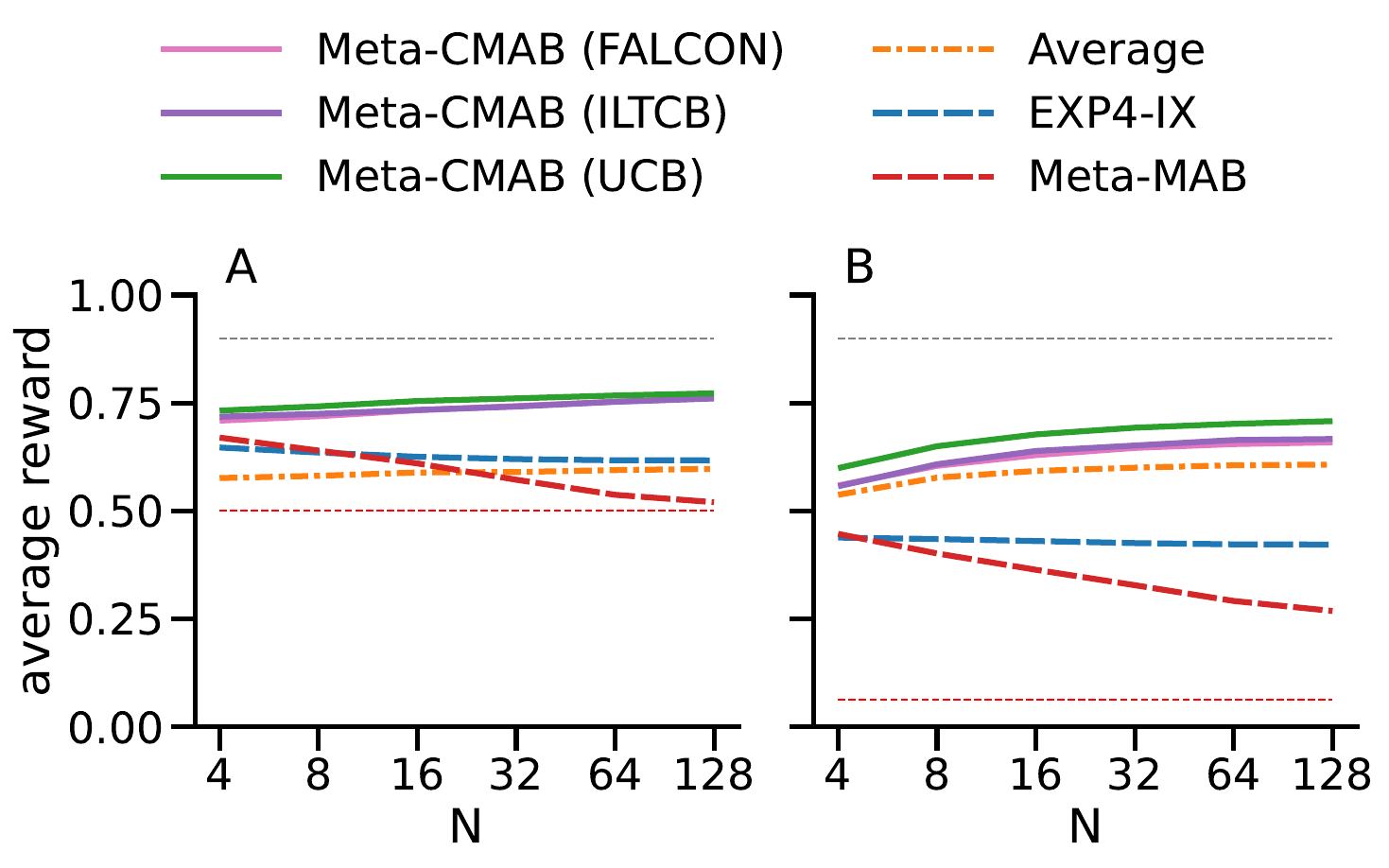}
     \end{subfigure}
     \hfill
     \begin{subfigure}[b]{0.48\textwidth}
     
\centering
 
 \includegraphics[width=1\textwidth]{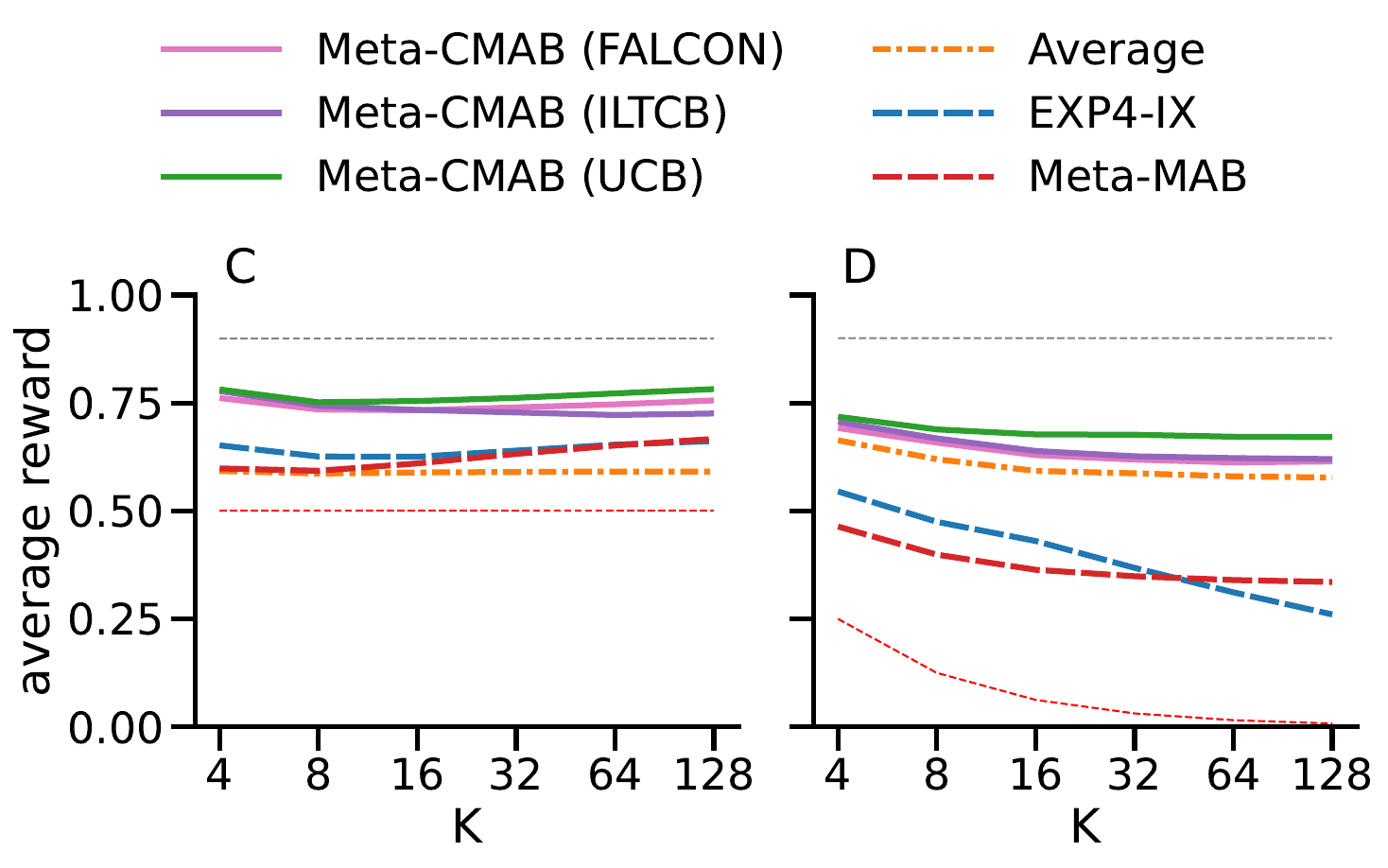}
     \end{subfigure}
     
        \caption{Average reward in function of number of experts (A and B) or arms (C and D) in the regression (A and C), or classification case (B and D). 
 The grey and red dashed lines mark the expected performance of respectively an optimal and random policy.}
        \label{fig:K_configs}
     \end{figure}

We evaluate the impact of increasing number of arms or experts on the performance of different algorithms. We group results by configuration in \autoref{fig:K_configs} and defer to supplementary figures \ref{fig:regr_KN_configs} and \ref{fig:class_KN_configs} for details grouped by expert distribution. In general, an increase in the number of experts (plots A and B of \autoref{fig:K_configs}) allows Meta-CMAB algorithms to improve their performance. 
  As our theoretical analysis shows (See \autoref{sec:misspecification}), an increased number of experts lowers the linear model's misspecification, which in turn induces enhanced performance for Meta-CMAB approaches. In contrast, Meta-MAB only updates the estimates of one expert each round. As a result, an increasing number of experts induces an increased need for exploration, which induces the observed performance loss. Finally, increasing the number of experts does not benefit EXP4-IX when the additional experts do not improve the performance of the single best expert. In which case increasing the number of experts degrades the performance polylogarithmically in the number of experts (see \autoref{tab:regret}), which explains the slight decrease in performance we observe.

In terms of arm counts, as we increase the number of arms (right-hand plots of \autoref{fig:K_configs}), the optimal policy in the regression case improves, as the reward of the best possible arm, drawn at random, increases with the number of arms. This improvement also induces an improvement in the performance of Meta-CMAB, Meta-MAB, and EXP4-IX. In the classification case, performance declines slightly as the number of arms increases. We observe that the decline is far more significant for EXP4-IX, and relate this to its regret bound which grows as $\sqrt{K}$ grows.

\subsection{Effect of changes in the number of trials} \label{sec:results_T}

\begin{figure}
\centering
 \includegraphics[width=.9\textwidth]{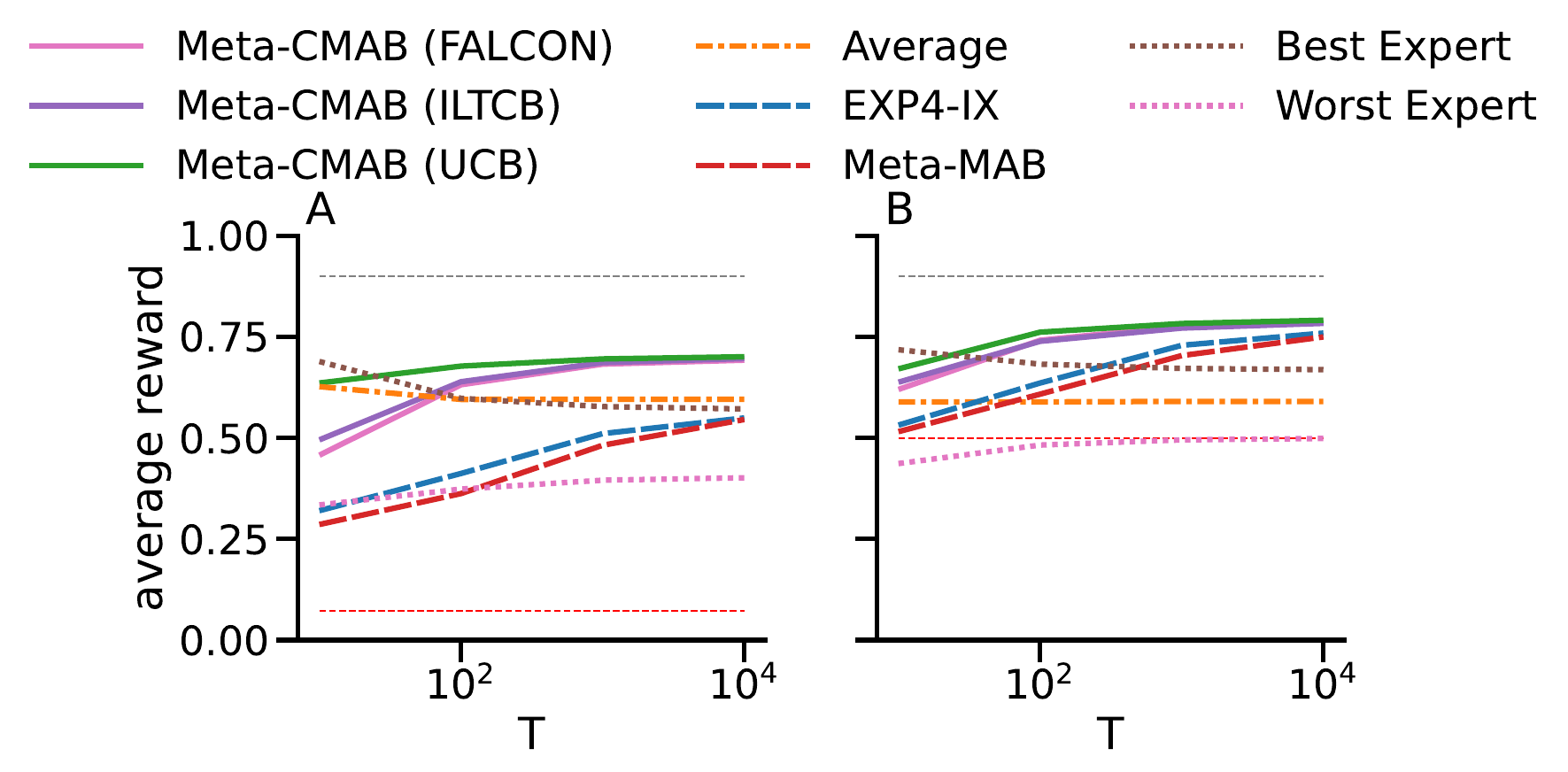}
 \caption{%
 Meta-CMAB compared to state-of-the-art on (A) \textbf{classification} or (B) \textbf{regression} bandits. Average reward in function of time horizon $T$. \textit{Best/Worst expert} is the performance obtained by the best/worst expert of each run. %
 The grey and red dashed lines mark the expected performance of respectively an optimal and random policy. %
 }\label{fig:trials}
\end{figure}

As the number of trials increases, the certainty about the optimal way to act on expert advice increases. However, given our interest in applying these methods to real-world decision making tasks wherein we cannot afford extensive training,
it is relevant to explore how these algorithms perform in function of the number of trials. 
We again provide an aggregated overview in \autoref{fig:trials} and defer to supplementary figures \ref{fig:regr_trials_errors_hom} to \ref{fig:class_trials_errors_het} for results broken down by $\epsilon$ and configuration.
This allows us to estimate the viability of the different algorithms when less training time is available. 

These figures confirm the intuition that a larger number of trials improves overall performance. It is noteworthy that the UCB variant already significantly outperforms the alternatives for $T=10$. 

What's more, on average the Meta-CMAB variants tend to surpass the best expert after $T=100$ trials. In contrast, the Meta-MAB approach can converge to high levels of performance, but performs essentially randomly for low values of $T$. As Meta-MAB only updates a single expert per timestep, its convergence is slower than EXP4-IX which updates its beliefs about all experts each timestep. For larger time horizons, Meta-MAB and EXP4-IX seem to converge towards Meta-CMAB's performance in the regression case.

\subsection{Validation on the crowdsourced datasets}
\begin{figure}

\centering

 \includegraphics[height=.26\textwidth]{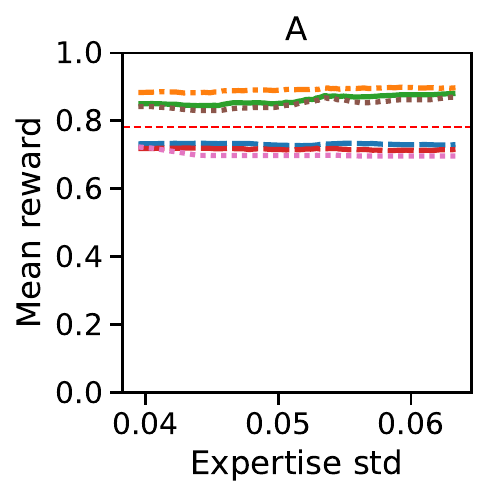}
 \includegraphics[height=.26\textwidth]{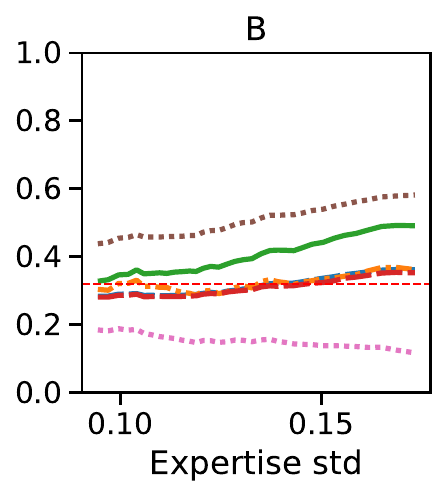}
 \includegraphics[height=.26\textwidth]{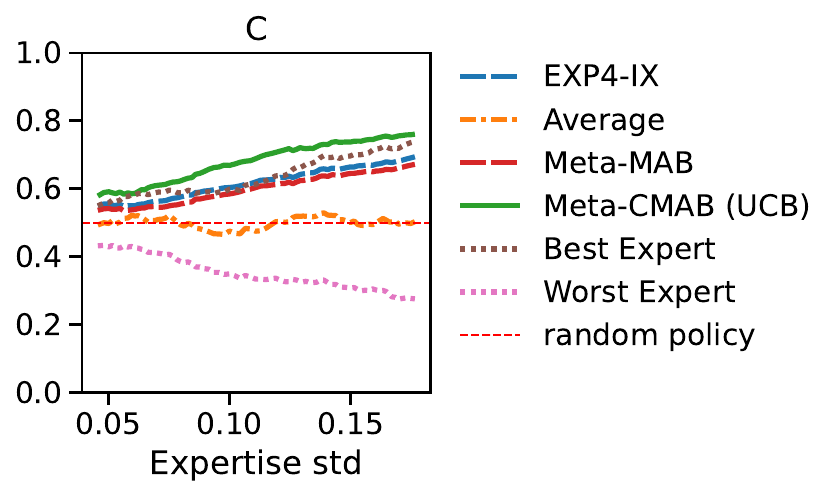}

 \caption{%
 Performance of different algorithms on human labeling datasets, for increasing levels of heterogeneity (as measured in terms of standard deviation in expert performance).  A) performance on the Emotion dataset, B) performance on the Recommendation dataset, and C) performance on the Duchenne dataset. \textit{Best/Worst expert} is the performance obtained by the best/worst expert of each run. %
 The red dashed line marks the expected performance of a random policy. Results are smoothed by applying a rolling average with window size $20$. 
 }\label{fig:human_datasets}
 
\end{figure}

Given that the results so far strongly favor the UCB variant of Meta-CMAB, we here restrict results to this variant.
\autoref{fig:human_datasets} compares the performance of our algorithms on various human datasets. 

For both the Duchenne and the Recommendation dataset, our Meta-CMAB algorithm significantly outperforms other algorithms. For the emotion dataset, our Meta-CMAB approach is only outperformed by the simple average approach. It is noteworthy that in this dataset, experts are relatively homogeneous and outperform a random policy. These are two conditions which allow an average to perform well. However, when these conditions do not hold, as is the case for the two other datasets, the performance of the average deteriorates significantly. In contrast, the Meta-CMAB performs well overall, and even surpasses the single best expert in $2$ out of $3$ datasets. In general the performance of the adaptive algorithms follows the performance of the single best expert. Specifically, as the best expert improves, adaptive algorithms improve. 

Comparing these results to those obtained with synthetic experts, the expert configuration for the Duchenne and Recommendation data set are more in line with the heterogeneous configuration (especially for large standard deviations), while experts for the Emotion data set tend to be homogeneous and better than random. In the latter case, and as was the case for the similar homogeneous configuration, the average performs strongly. In all other cases, i.e., when the optimal weighing of experts differs significantly from an average, the Meta-CMAB approach performs best. Therefore, if experts are expected to differ significantly in terms of expertise, or if worse-than-random experts are likely, the Meta-CMAB approach should be preferred.

\section{Conclusion}
In this paper we explored the presence of systematic biases in expert advice and how they affect the performance of several CDM algorithms under different expert configurations, and for both human and synthetic experts.  In particular, we found that our proposed Meta-CMAB approach can implicitly identify such biases and counteract them, provided there is a correlation (whether positive or negative) between the experts' advice and the outcomes. In contrast, previous approaches to bandits with expert advice, such as EXP4-IX, and Meta-MAB, are constrained by the performance of the single best expert and as a result these algorithms rapidly degrade in performance when biases are present. 

Our exploration of different expert configurations demonstrates the potential for Meta-CMAB on a wide variety of collective decision-making tasks, provided an objective feedback from decisions (i.e., a reward) is available. Our first results showed how a group of experts can benefit from the Meta-CMAB algorithm. In particular, Meta-CMAB is able to exploit the collective, whereas EXP4-IX, and Meta-MAB tend to converge towards the exploitation of a single expert. Meta-CMAB thus really reaps the benefits of consulting multiple experts as it finds solutions better than what can be provided by the best expert in the group. 

We subsequently validated the improvements provided by Meta-CMAB on human expertise derived from crowd-sourced tasks. In particular, we found that Meta-CMAB surpasses previous adaptive approaches (EXP4-IX, Meta-MAB) on all problems with human expertise. We also found that, only when experts are homogeneous and better-than-random --- the ideal configuration for an average --- is Meta-CMAB slightly outperformed by a simple average. In all other configurations however, Meta-CMAB is a significant improvement over the simple average.

Of further interest is the observation that Meta-CMAB is able to beneficially exploit advice provided by weak experts. This opens the way to applications wherein imperfect advice is prevalent, for example through crowd-sourcing, wherein a heterogeneous configuration of experts is likely. In such scenarios we found that Meta-CMAB assigns an importance to experts which is directly correlated to their expected performance. Similarly, we also demonstrated that Meta-CMAB achieves high performance when experts are moderately polarized (i.e., when two groups of experts with opposing advice occur), which could for example be beneficial in policy-making, wherein polarization is a recurring nefarious phenomenon. 

Finally, we showed that Meta-CMAB not only outperforms other algorithms over time but that it does so by quickly achieving high performance. Consequently, Meta-CMAB is highly useful for CDM applications wherein only a small number of testing steps can be executed before really being deployed. 

It would be worthwhile to explore now how Meta-CMAB could be used in existing applications which make use of Average aggregation or EXP4-IX. \cite{clement2015multi} for example leverage EXP4 to identify appropriate activities in an intelligent tutoring system. Similarly, the use of Average aggregation has been explored in medical diagnostics, \cite{kammer2017potential} for example enhance diagnostics by computing a (weighted) average of human experts for a simulated diagnostic problem. Similar research has found that combining the opinion of a group of radiologists can improve mammography screening beyond the single best expert \citep{wolf2015collective}. The (weighted) averages in these approaches could be replaced by Meta-CMAB to guard for biased expertise. 

To conclude, biases are not exclusive to human experts. As machine learning efforts in medical diagnostics have shown, deep learning models can struggle to generalize from one hospital to another \citep{zech2018variable}. In such scenarios too Meta-CMAB could be applied by framing the diagnostic problem as a bandit and using models trained on other hospitals as experts.  As the COVID-19 pandemic has highlighted, conflicting expertise is inevitable, and, in the absence of prior knowledge to filter expertise, our Meta-CMAB method offers a promising approach to re-conciliate and exploit the knowledge of biased as well as conflicting experts.

\section*{Acknowledgments} A.A. is supported by a FRIA grant by the National Fund for Scientific Research (F.N.R.S.) of Belgium. T.L. is supported by the F.N.R.S. project with grant numbers 31257234 and 40007793, the F.W.O. project with grant nr. G.0391.13N, the Service Public de Wallonie Recherche under grant n\textdegree 2010235–ARIAC by DigitalWallonia4.ai. T.L and A.N. benefit from the support of the Flemish Government through the AI Research Program. T.L., V.T and A.N. acknowledge the support by TAILOR, a project funded by EU Horizon 2020
research and innovation program under GA No 952215. A.N. and T.L. also acknowledge the support by the Flemish Government through the AI Research Program. The resources and services used in this work were provided by the VSC (Flemish Supercomputer Center), funded by the Research Foundation - Flanders (FWO) and the Flemish Government.

\newpage

\bibliographystyle{elsarticle-harv} 
\bibliography{main}

\appendix 

\renewcommand{\thefigure}{S.\arabic{figure}}
\renewcommand\thefigure{S.\arabic{figure}} 
\renewcommand\thesection{S.\arabic{section}} 
\renewcommand\thealgorithm{S.\arabic{algorithm}} 
\setcounter{figure}{0}
\newpage

\begin{center}
{\LARGE\bfseries Supplementary Information}
\end{center}

\section{Meta-MAB}\label{sec:metamab}
We solve the Meta-MAB using Thompson Sampling \citep{ThompsonONTL} and use a Beta distribution to model the priors. 
\begin{algorithm}[ht]
 \caption{Description of the Meta-MAB algorithm}
 \label{alg:metamab}
 \begin{algorithmic}[1]
 \State Set $\vv{\alpha}_{1} = \vec{1}$ and $\vv{\beta}_{1} = \vec{1}$ each of size $N$.
 \For{$t=1, 2, ..., T$}
 \State {Get expert advice $\pmb{ \tilde{f}}_t=\{\vv{\tilde{f}}^1_t, ..., \vv{\tilde{f}}^N_t\}$ } \label{step:outline:advice}%
 \For{$n=1, 2, ..., N$} 
 \State Sample $\hat{\theta}_n \sim \beta(\alpha^n_t,\beta^n_t )$
 \EndFor
 \State $n_t = \argmax_{n=1}^N \hat{\theta}_n$ // Choose expert
 \State Draw arm $k_t$ according to ${\vv {\tilde{f}}^{n_t}_t}$, and receive reward $r_{t}$.
 \State $\triangleright$ Update beliefs about chosen expert
 \State $\alpha^{n_t}_{t+1} = \alpha^{n_t}_t + r_t$
 \State $\beta^{n_t}_{t+1} = \beta^{n_t}_t + (1-r_t)$
 \EndFor
 \end{algorithmic}
 \end{algorithm}

\section{EXP4-IX}\label{sec:exp4pcon}
The following algorithm adapts EXP4-IX \citep{neu2015explore} for regressor advice by using the greedy policies induced by the advice. Let $\vv{\tilde{f}}^n_t$ be the regressor advice of expert $n$, we denote by $\pi({\tilde{f}^n})$ the probability distribution concentrated on the arms with maximal value estimates, and by $\pi_k({\tilde{f}^n})$ the probability for arm $k$ induced by $\tilde{f}^n$.  
 \begin{algorithm}[ht]
 \caption{EXP4-IX}
 \label{alg:exp4p}
 \begin{algorithmic}[1]
 \Require $\delta > 0$, $M \ge 0$
 \State Define $\gamma=\sqrt{2\frac{\ln N}{KT}}$, set $\vv{w}_{1} = \vec{1}$ of size $N$.
 \For{$t=1, 2, ..., T$}
 \State {Get expert advice $\pmb{ \tilde{f}}_t=\{\vv{\tilde{f}}^1_t, ..., \vv{\tilde{f}}^N_t\}$, } %
 \For{$k=1, 2, ..., K$} \quad $\triangleright$ compute weighted average 
 \State \hspace{7em}$ p_{k,t} = \sum_{n=1}^N \frac{exp({ w}^n_t)}{\sum_{n'=1}^N exp({ w}^{n'}_t)} \pi_k({\tilde{f}^n})$ \label{alg:exp4p-aggregation} %
 \EndFor
 \State Draw arm $k_t$ according to ${\vv p}_t$, and receive reward $r_{t}$.
 \For{$n=1, ..., N$} \quad $\triangleright$ Update weights
 \begin{align*} 
 \hat{y}^n_{t} &= \frac{\pi_k({\tilde{f}^n}) {r}_{t}}{p_{k_t,t}+\gamma} \\
 w^n_{t+1} &= w^n_{t} +2\gamma\hat{y}^n_{t} 
 \end{align*}
 \EndFor
 \EndFor
 \end{algorithmic}
 \end{algorithm}

\begin{figure}
\centering
 
 \includegraphics[width=1\textwidth]{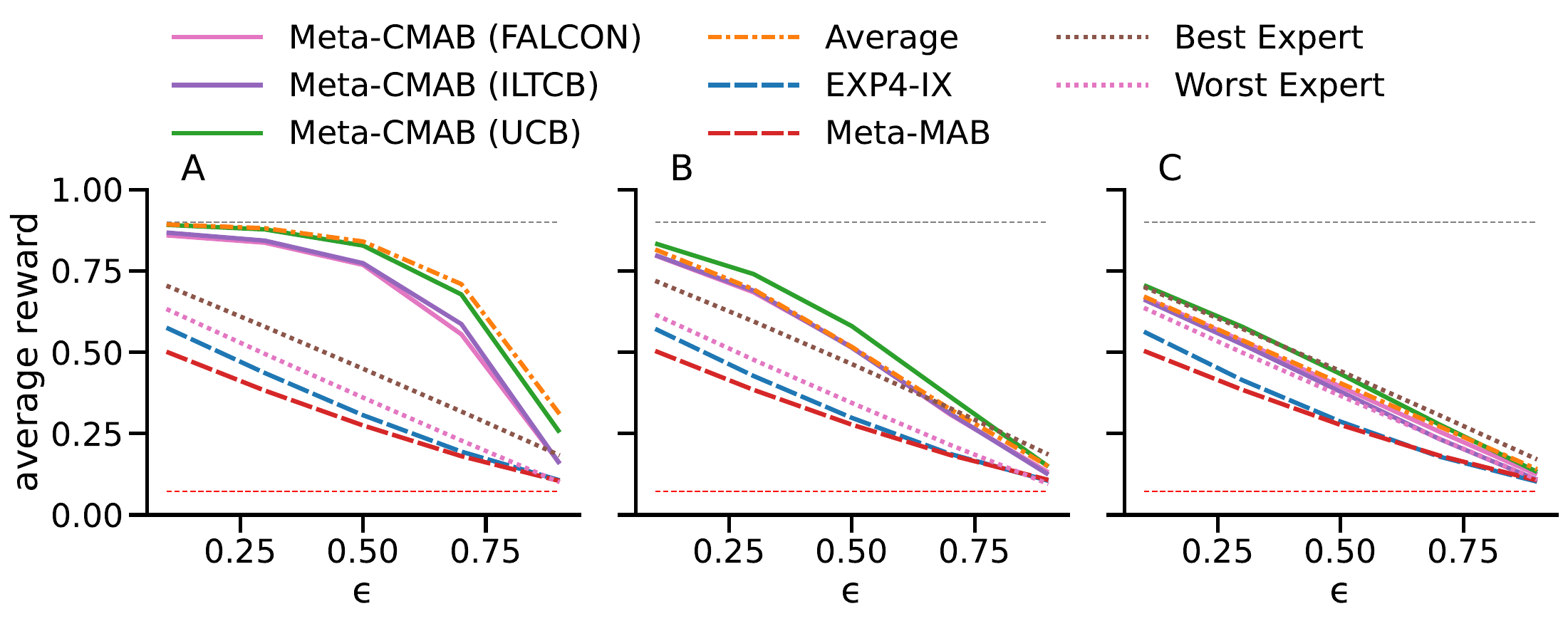}

 \caption{%
 \textbf{Homogeneous} experts, Meta-CMAB compared to state-of-the-art on \textbf{classification} bandits. Average reward in function of expert misspecification (i.e., $\epsilon$) for different levels of \textbf{polarization}; A) low polarization ($\rho=0.1)$), B) medium polarization ($\rho=0.5)$), and C) high polarization ($\rho=0.9)$). \textit{Best/Worst expert} is the performance obtained by the best/worst expert of each run. %
 The grey and red dashed lines mark the expected performance of respectively an optimal and random policy. %
 }\label{fig:class_correlation_error_hom}
\end{figure}

\begin{figure}
\centering
 
 \includegraphics[width=1\textwidth]{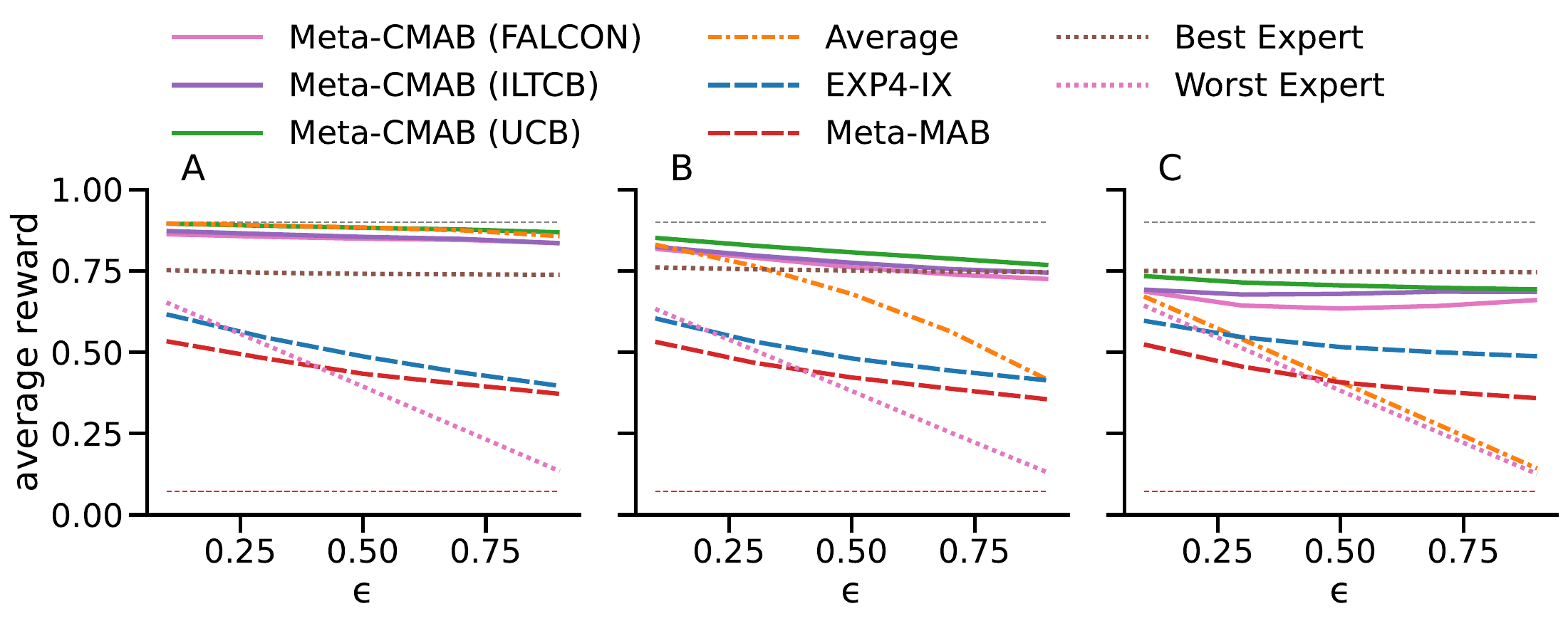}

 \caption{%
 \textbf{Heterogeneous} experts, Meta-CMAB compared to state-of-the-art on \textbf{classification} bandits. Average reward in function of expert misspecification (i.e., $\epsilon$) for different levels of \textbf{polarization}; A) low polarization ($\rho=0.1)$), B) medium polarization ($\rho=0.5)$), and C) high polarization ($\rho=0.9)$). \textit{Best/Worst expert} is the performance obtained by the best/worst expert of each run. %
 The grey and red dashed lines mark the expected performance of respectively an optimal and random policy. %
 }\label{fig:class_correlation_error_het}
\end{figure}

\begin{figure}
\centering
 
 \includegraphics[width=1\textwidth]{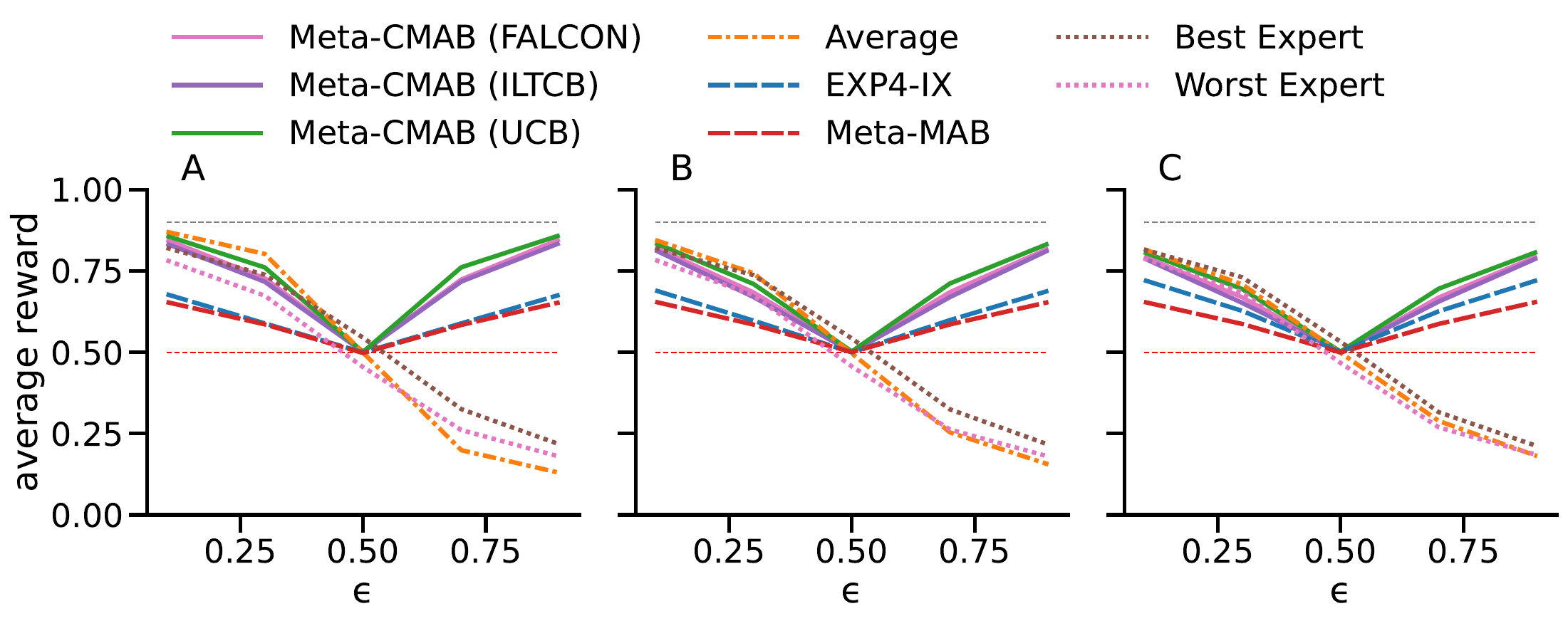}

 \caption{%
  \textbf{Homogeneous} experts, Meta-CMAB compared to state-of-the-art on \textbf{regression} bandits. Average reward in function of expert misspecification (i.e., $\epsilon$) for different levels of  \textbf{polarization}; A) low polarization ($\rho=0.1)$), B) medium polarization ($\rho=0.5)$), and C) high polarization ($\rho=0.9)$). \textit{Best/Worst expert} is the performance obtained by the best/worst expert of each run. %
 The grey and red dashed lines mark the expected performance of respectively an optimal and random policy. %
 }\label{fig:regr_correlation_error_hom}
\end{figure}

\begin{figure}
\centering
 
 \includegraphics[width=1\textwidth]{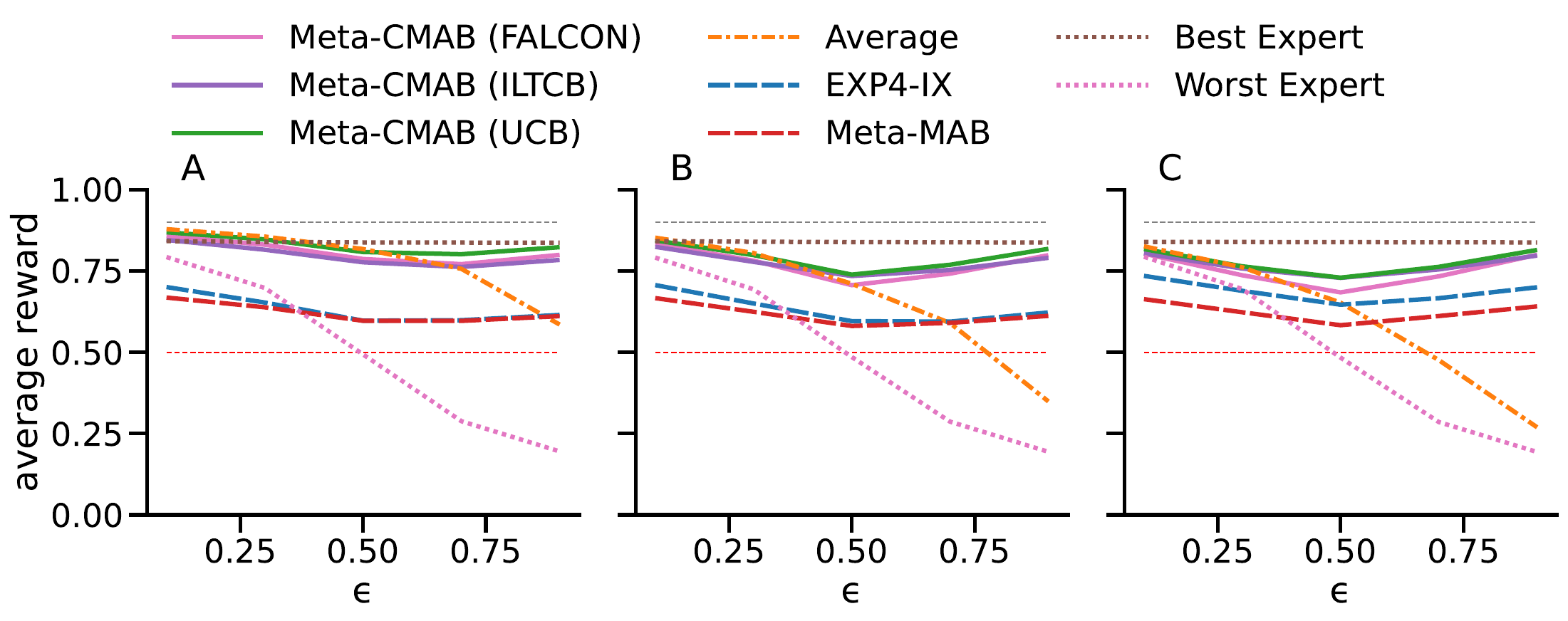}

 \caption{%
\textbf{Heterogeneous} experts, Meta-CMAB compared to state-of-the-art on \textbf{regression} bandits. Average reward in function of expert misspecification (i.e., $\epsilon$) for different levels of  \textbf{polarization}; A) low polarization ($\rho=0.1)$), B) medium polarization ($\rho=0.5)$), and C) high polarization ($\rho=0.9)$). \textit{Best/Worst expert} is the performance obtained by the best/worst expert of each run. %
 The grey and red dashed lines mark the expected performance of respectively an optimal and random policy. %
 }\label{fig:regr_correlation_error_het}
\end{figure}

\begin{figure}

     \begin{subfigure}[b]{0.48\textwidth}
\centering
 
 \includegraphics[width=1\textwidth]{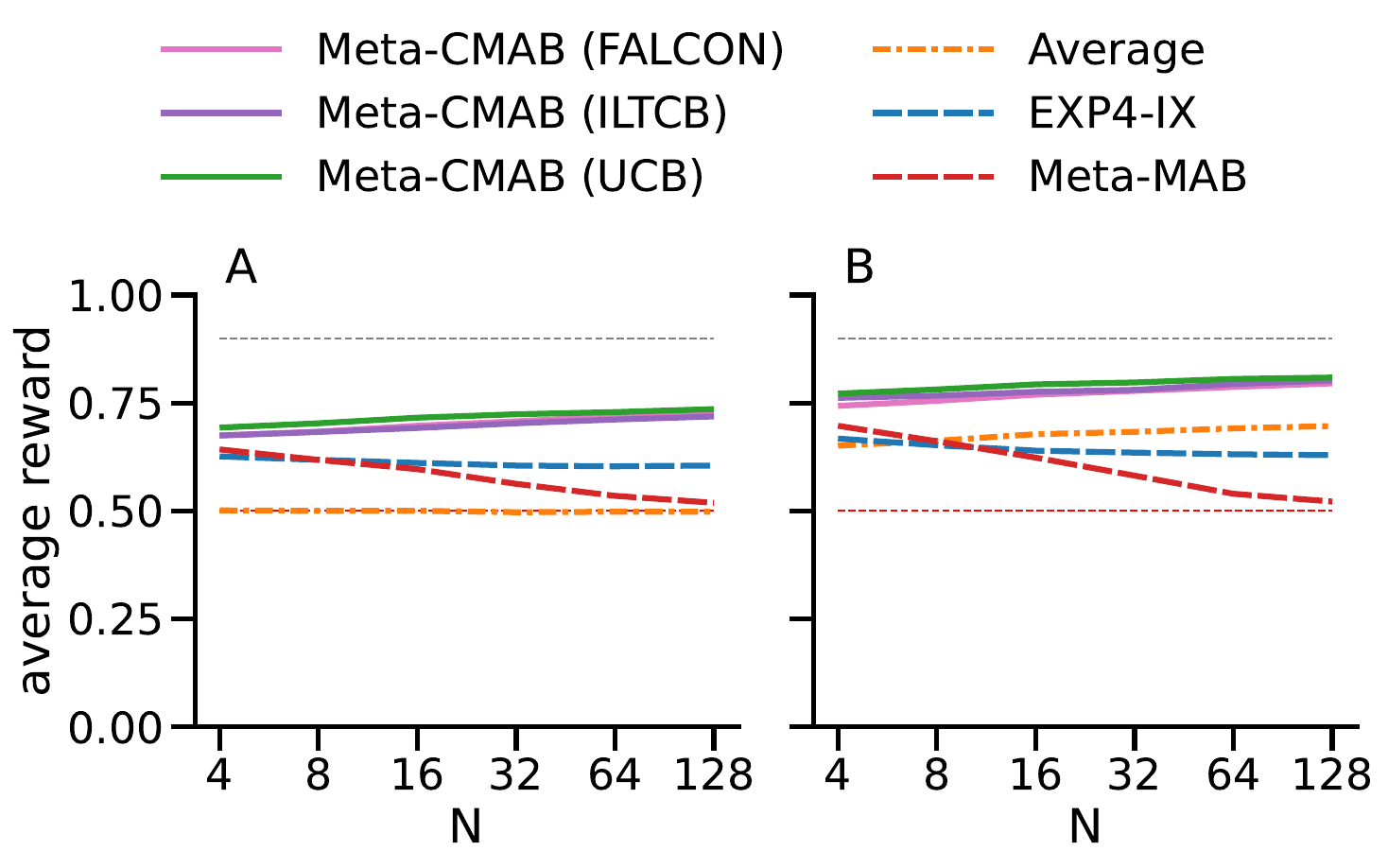}
\label{fig:regr_config_N}
     \end{subfigure}
     \hfill
     \begin{subfigure}[b]{0.48\textwidth}
     
\centering
 
 \includegraphics[width=1\textwidth]{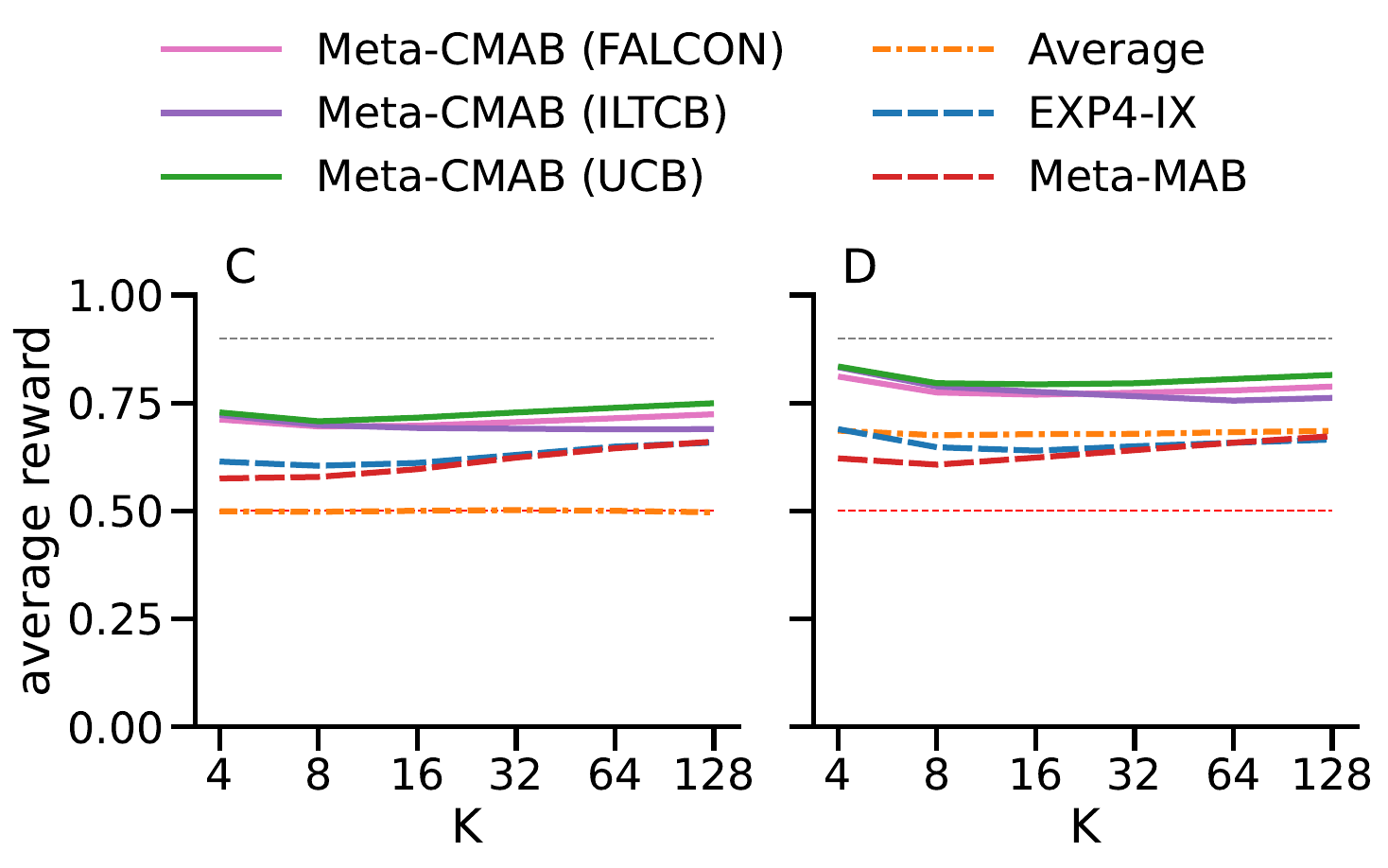}
\label{fig:regr_config_K}
     \end{subfigure}
     
        \caption{Meta-CMAB compared to state-of-the-art on \textbf{regression} bandits. Average reward in function of number of experts (left) or arms (right) in the A) homogeneous, or B) heterogeneous case. 
 The grey and red dashed lines mark the expected performance of respectively an optimal and random policy.}
        \label{fig:regr_KN_configs}
     \end{figure}

\begin{figure}

     \begin{subfigure}[b]{0.48\textwidth}
\centering
 
 \includegraphics[width=1\textwidth]{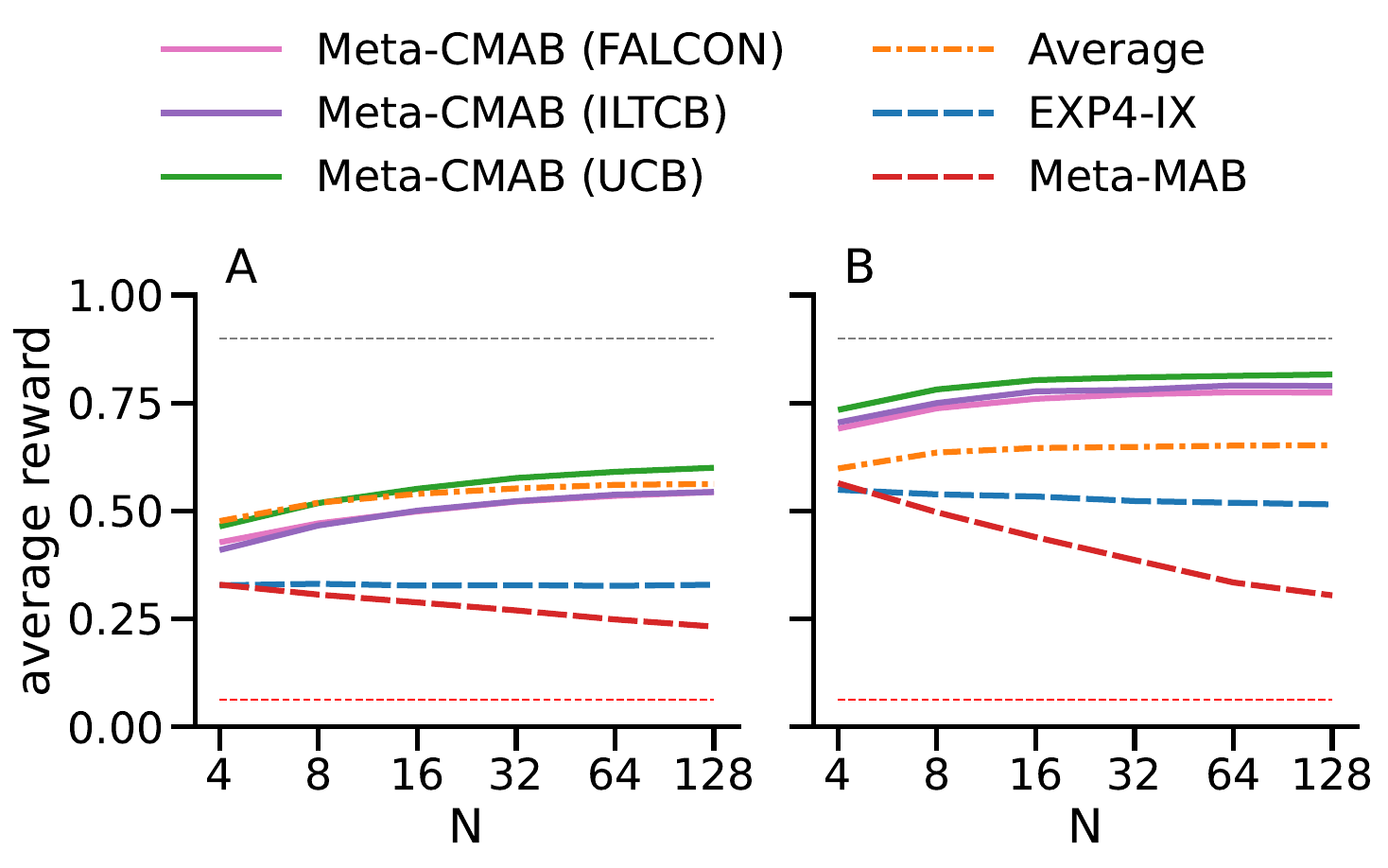}
     \end{subfigure}
     \hfill
     \begin{subfigure}[b]{0.48\textwidth}
     
\centering
 
 \includegraphics[width=1\textwidth]{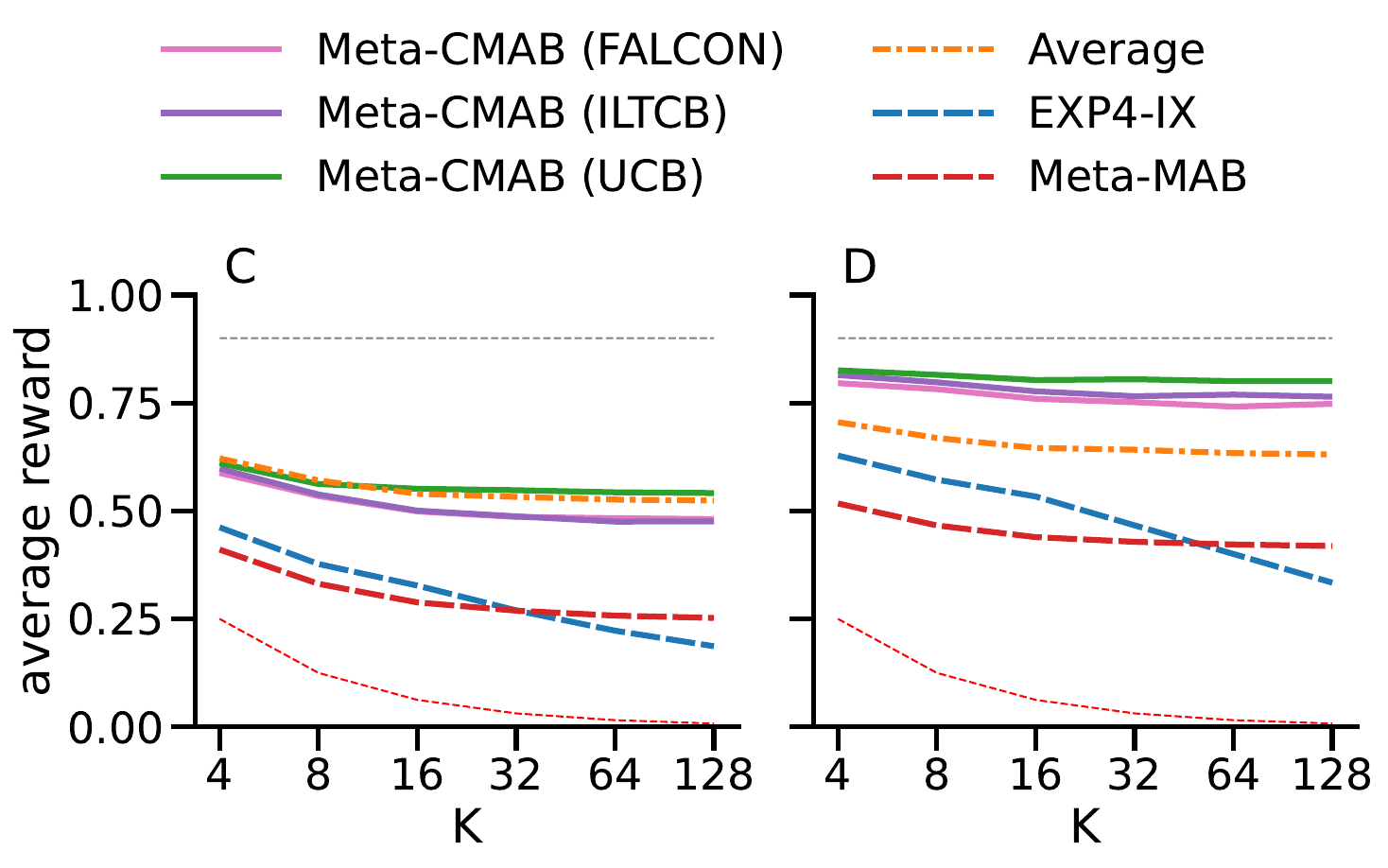}
     \end{subfigure}
     
        \caption{Meta-CMAB compared to state-of-the-art on \textbf{classification} bandits. Average reward in function of number of experts (left) or arms (right) in the A) homogeneous, or B) heterogeneous case. 
 The grey and red dashed lines mark the expected performance of respectively an optimal and random policy.}
        \label{fig:class_KN_configs}
     \end{figure}

\begin{figure}
\centering
 
 \includegraphics[width=1\textwidth]{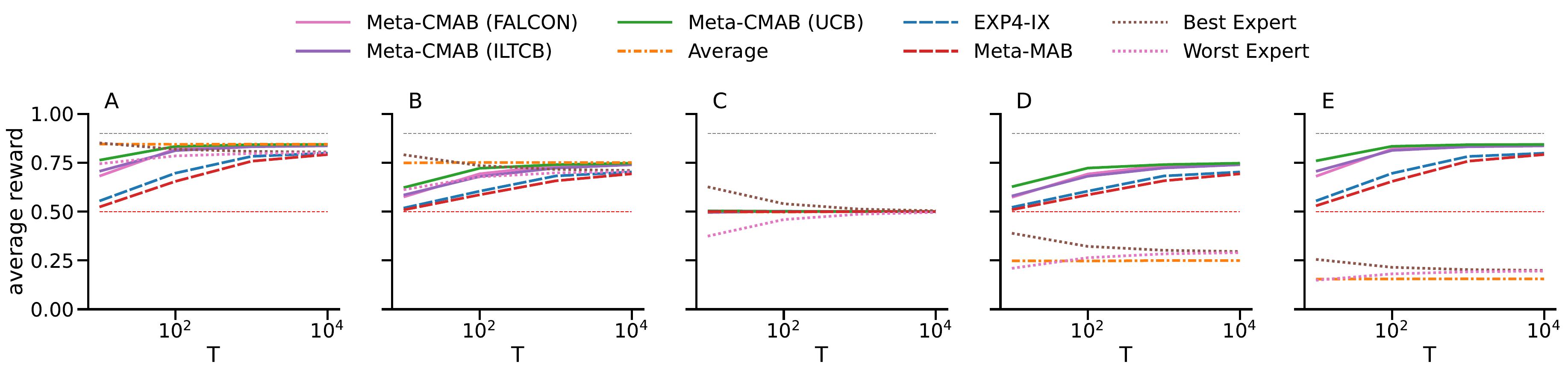}

 \caption{%
 \textbf{Homogeneous} experts, Meta-CMAB compared to state-of-the-art on \textbf{regression} bandits.  Average reward in function of expert misspecification ($\epsilon$)  for increasing number of \textbf{trials}; A) $\epsilon=0.1$, B) $\epsilon=0.3$, C) $\epsilon=0.5$, D) $\epsilon=0.7$, and E) $\epsilon=0.9$. \textit{Best/Worst expert} is the performance obtained by the best/worst expert of each run. %
 The grey and red dashed lines mark the expected performance of respectively an optimal and random policy. %
 }\label{fig:regr_trials_errors_hom}
\end{figure}

\begin{figure}
\centering
 
 \includegraphics[width=1\textwidth]{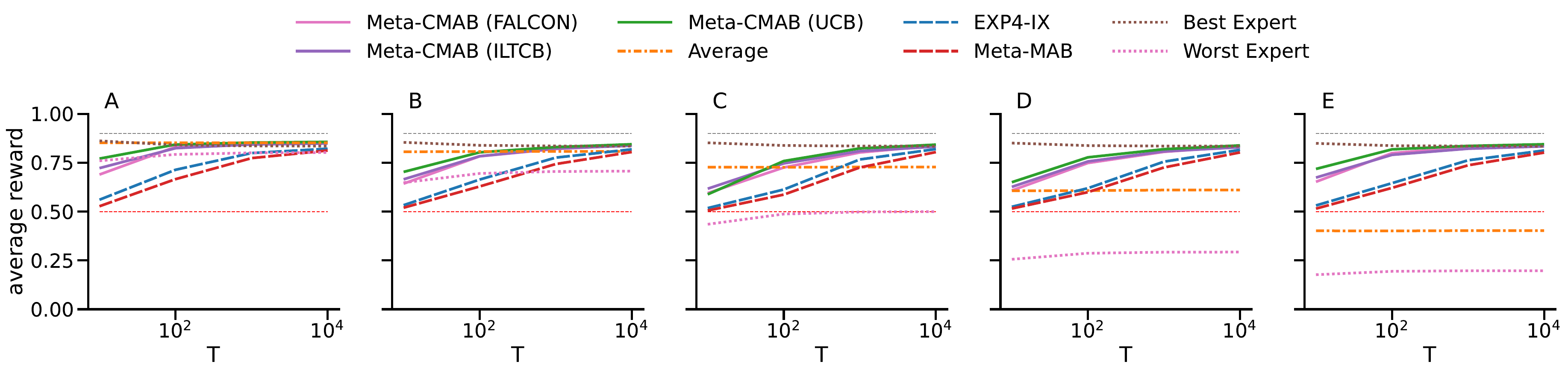}

 \caption{%
 \textbf{Heterogeneous} experts, Meta-CMAB compared to state-of-the-art on \textbf{regression} bandits. Average reward in function of expert misspecification ($\epsilon$) for increasing number of \textbf{trials};  A) $\epsilon=0.1$, B) $\epsilon=0.3$, C) $\epsilon=0.5$, D) $\epsilon=0.7$, and E) $\epsilon=0.9$. \textit{Best/Worst expert} is the performance obtained by the best/worst expert of each run. %
 The grey and red dashed lines mark the expected performance of respectively an optimal and random policy. %
 }\label{fig:regr_trials_errors_het}
\end{figure}
\begin{figure}
\centering
 
 \includegraphics[width=1\textwidth]{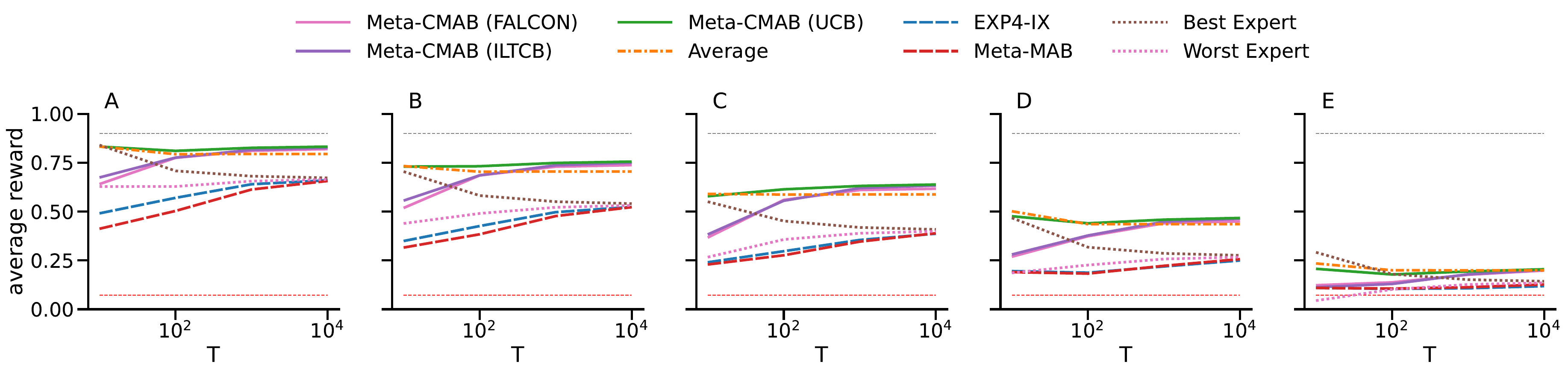}

 \caption{%
 \textbf{Homogeneous} experts, Meta-CMAB compared to state-of-the-art on \textbf{classification} bandits. Average reward in function of expert misspecification ($\epsilon$)  for increasing number of \textbf{trials}; A) $\epsilon=0.1$, B) $\epsilon=0.3$, C) $\epsilon=0.5$, D) $\epsilon=0.7$, and E) $\epsilon=0.9$. \textit{Best/Worst expert} is the performance obtained by the best/worst expert of each run. %
 The grey and red dashed lines mark the expected performance of respectively an optimal and random policy. %
 }\label{fig:class_trials_errors_hom}
\end{figure}
\begin{figure}
\centering
 
 \includegraphics[width=1\textwidth]{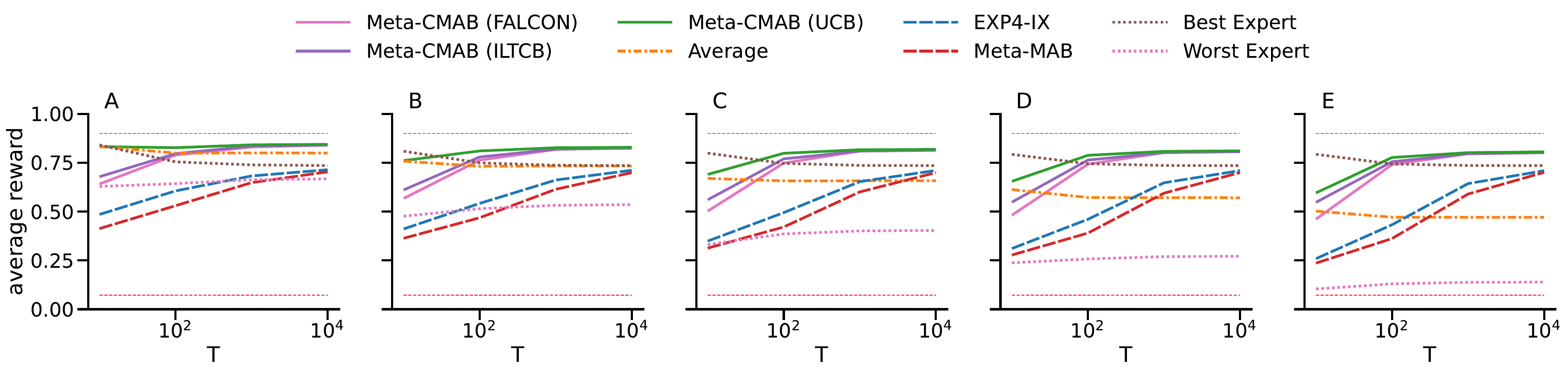}

 \caption{%
 \textbf{Heterogeneous} experts, Meta-CMAB compared to state-of-the-art on \textbf{classification} bandits. Average reward in function of expert misspecification ($\epsilon$)  for increasing number of \textbf{trials}; A) $\epsilon=0.1$, B) $\epsilon=0.3$, C) $\epsilon=0.5$, D) $\epsilon=0.7$, and E) $\epsilon=0.9$. \textit{Best/Worst expert} is the performance obtained by the best/worst expert of each run. %
 The grey and red dashed lines mark the expected performance of respectively an optimal and random policy. %
 }\label{fig:class_trials_errors_het}
\end{figure}

\end{document}